\documentclass[twoside,11pt]{article}
\usepackage[preprint]{jmlr2e}

\makeatletter

\let\proof\relax        
\let\c@theorem\relax
\makeatother

\usepackage{amsmath,amsthm,mathtools}
\AtBeginDocument{}
\usepackage{dsfont}
\usepackage{enumitem}
\usepackage{booktabs}
\usepackage{tabularx}
\usepackage{xcolor}
\usepackage{microtype}
\usepackage{lastpage}

\hypersetup{
  hidelinks,
  pdftitle={A Law of Robustness for Log-Concave Covariates},
  pdfauthor={Tobias J. Peter},
  pdfkeywords={law of robustness, exponential concentration, log-concave measures,
    Poincare inequality, Lipschitz interpolation}
}

\ShortHeadings{A Law of Robustness for Log-Concave Covariates}{Peter}
\jmlrheading{}{2026}{}{}{}{}{Tobias J. Peter}
\firstpageno{1}

\theoremstyle{definition}
\newtheorem{theorem}{Theorem}[section]
\newtheorem{lemma}[theorem]{Lemma}
\newtheorem{proposition}[theorem]{Proposition}
\newtheorem{corollary}[theorem]{Corollary}
\newtheorem{fact}[theorem]{Fact}
\newtheorem{definition}[theorem]{Definition}
\newtheorem{assumption}[theorem]{Assumption}
\newtheorem{example}[theorem]{Example}
\theoremstyle{remark}
\newtheorem{remark}[theorem]{Remark}

\numberwithin{table}{section}

\newcommand{\R}{\mathbb{R}}
\newcommand{\Id}{\mathds{1}} 
\newcommand{\E}{\mathbb{E}}
\newcommand{\Prob}{\mathbb{P}}
\newcommand{\Var}{\mathrm{Var}}
\newcommand{\Lip}{\mathrm{Lip}}
\newcommand{\LF}{\mathrm{Lip}_{F}}
\newcommand{\eps}{\varepsilon}

\newcommand{\cF}{\mathcal{F}}
\newcommand{\cW}{\mathcal{W}}
\newcommand{\cN}{\mathcal{N}}
\newcommand{\cG}{\mathcal{G}}
\newcommand{\cA}{\mathcal{A}}
\newcommand{\cC}{\mathcal{C}}
\newcommand{\inner}[2]{\langle #1,\,#2\rangle}
\newcommand{\norm}[1]{\lVert #1\rVert}
\newcommand{\expTail}{\operatorname{expTail}}

\title{\bf A Law of Robustness for Log-Concave Covariates}
\author{\name Tobias J. Peter \email tobias.peter@zhaw.ch\\
       \addr ZHAW Zurich University of Applied Sciences\\
       Institute of Computational Life Sciences, Schloss, 8820 Wädenswil}
\begin{document}
\editor{}
\maketitle

\begin{abstract}%
Bubeck and Sellke's universal law of robustness says that fitting $n$ noisy
samples below the noise level with a class of $p$ real parameters forces a
Lipschitz constant of order $\sqrt{nd/p}$: smooth interpolation
costs a factor~$d$ of overparametrization. Its proof
assumes Gaussian-type (isoperimetric) concentration of the covariates. 
We show that exponential concentration suffices: the same threshold,
with the same absolute constant, holds when $Q\le61\,n$, where $Q$
measures parameter-net complexity and includes a confidence term.
Without this condition the law still holds, with an extra
factor of order $\sqrt{n/Q}$ when $Q\ge n$; for $Q\gg n$ this loss
occurs for some normalized log-concave covariates. For these covariates, Klartag's bound on the
KLS constant yields the result with a $\sqrt{\log d}$ loss in the
Lipschitz lower bound, which a recently announced proof of the KLS
conjecture removes. The argument covers $\psi_\alpha$ tails for every
$\alpha\in[1,2]$.
\end{abstract}


\section{Introduction}\label{sec:intro}

Modern deep networks often use far more parameters than training equations. \citet{BS21}
explain why: \emph{smooth} interpolation is more costly than interpolation alone. Their
\emph{universal law of robustness} asserts that, for a $p$-parameter class and a
$c$-isoperimetric covariate distribution on $\R^d$ (Definition~\ref{def:isop} below),
any $f$ in the class fitting $n$ noisy samples $\eps$ below the noise level must satisfy
\begin{equation}\label{eq:BSlaw}
 \Lip(f)\;\gtrsim\;\eps\sqrt{\frac{nd}{p}}\,,
\end{equation}
up to logarithmic factors. A uniformly Lipschitz fit therefore requires $p\gtrsim nd$
parameters.

The Gaussian-type concentration required by this law excludes natural covariate
distributions. Normalized log-concave covariates satisfy a Poincar\'e inequality, which
guarantees exponential concentration but need not yield Gaussian-type tails. A product of
Laplace laws, for example, is not $c$-isoperimetric for any finite $c$
(Proposition~\ref{rem:lcnoconverse}). We show that exponential concentration suffices for
the same Lipschitz lower bound, subject to an additional capacity condition.

Our proof builds on the noise-correlation reduction and parameter-net argument of
\citet{BS21}. We formulate this reduction as a master theorem whose probabilistic
conditions can be verified under different concentration hypotheses. To obtain the
law under exponential concentration, we develop a bounded-range product-MGF estimate
and control the resulting two-branch tail uniformly over the parameter net. The main
theorem retains the baseline Lipschitz threshold under this weaker concentration
hypothesis. Without the capacity condition, the threshold drops by at most a factor of
order $\sqrt{n/Q}$. A separate construction shows that for $Q\gg n$ this drop occurs for
some normalized log-concave covariates.

The new ingredient is the bound on the noise--fluctuation pairing. Under exponential
concentration, its moment-generating function admits a quadratic bound at the variance
scale only over a bounded range of the Chernoff parameter (Lemma~\ref{lem:lckey}). This
restriction is intrinsic (Remark~\ref{rem:lccap}) and leads to a tail bound with quadratic
and linear regimes (Corollary~\ref{cor:lcpairing}). The additional capacity condition
controls $Q/n$, where $Q$ is the complexity term defined in \eqref{eq:Q} below, and
ensures that the linear regime is strong enough for the union
bound.

\subsection{Main Results}\label{sec:mainresults}

Throughout, write
\begin{equation}\label{eq:Q}
 Q \;\coloneqq\; p\,\log\!\Big(\tfrac{48\,WJ}{\eps}\Big) \;+\; \log\!\Big(\tfrac{16}{\delta}\Big)
 \qquad\text{and}\qquad
 L_0 \;\coloneqq\; \frac{\eps}{2^{9}}\sqrt{nd/Q}\,,
\end{equation}
with $\eps$ the overfitting margin, $\delta$ the failure probability, and $p$, $W$, $J$
the parameter count, diameter and parameter-Lipschitz constant of the class
(Definition~\ref{def:param}); Sections~\ref{sec:prelim} and~\ref{sec:skeleton} define the
model and the parametrization precisely. 
Informally (precise statements in the indicated theorems):

\begin{enumerate}[label=\textbf{(\Alph*)},leftmargin=2.2em]
\item \textbf{Exponential concentration (Theorem~\ref{thm:logconcave},
Propositions~\ref{prop:lcburnin} and~\ref{prop:lcbeyond}).}
Isoperimetry can be relaxed to \emph{$c$-exponential} concentration
(Definition~\ref{def:expconc}):
the law holds at the unchanged rate \eqref{eq:BSlaw} and absolute constant,
at the price of the burn-in $Q\le61\,n$. Without the burn-in, the law holds at the level
$\frac{\eps}{2^6}\sqrt{d/c}\,\frac{n}{n+Q}$ (Proposition~\ref{prop:lcbeyond}), which for
$Q\gg n$ is smaller than the rate \eqref{eq:BSlaw} by a factor of order $\sqrt{n/Q}$.
In the worst case this loss is real: for $Q>2^{28}n$, the construction of
Proposition~\ref{prop:lcburnin} matches the degraded level up to an absolute factor.

\item \textbf{Log-concave covariates (Corollary~\ref{cor:logconcave}).}
By Klartag's bound on the KLS constant,
Corollary~\ref{cor:logconcave} yields the law for every full-dimensional log-concave $\mu$
satisfying $\norm{\mathrm{Cov}(\mu)}_{\mathrm{op}}\le\kappa/d$, with $c=O(\kappa\log d)$.
The recently announced proof of the KLS conjecture \citep[Theorem~1.1]{BKL26} removes the
$\log d$, giving $c=O(\kappa)$; we state this as a separate part of the corollary, since
that proof has not yet been refereed (Remark~\ref{rem:kls}). The burn-in of
Theorem~\ref{thm:logconcave} remains necessary within this class: the witness of Proposition~\ref{prop:lcburnin} is a product of Laplace
laws with covariance $\frac2d\Id_d$.

\item \textbf{The $\psi_\alpha$ scale
(Corollary~\ref{rem:lcpsialpha}, Proposition~\ref{prop:poincexp}).}
Theorem~\ref{thm:logconcave} holds verbatim under $\psi_\alpha$ concentration for every
$\alpha\in[1,2]$. We use the classical implication from a Poincar\'e inequality to
exponential concentration \citep{GM83,BL97},
with the explicit constant $9c_0$ for a $d$-normalized Poincar\'e inequality.
\end{enumerate}

\noindent
Table~\ref{tab:results} compares the conclusions and finite-sample conditions.

\begin{table}[t]
\centering
{\small
\setlength{\tabcolsep}{4pt}
\renewcommand{\arraystretch}{1.15}
\begin{tabularx}{\textwidth}{@{}l>{\raggedright\arraybackslash}Xl>{\raggedright\arraybackslash}p{0.28\textwidth}@{}}
\toprule
 & covariate hypothesis & lower bound & sample conditions\\
\midrule
Thm.~\ref{thm:baseline} & $c$-isoperimetry (baseline) & $L_0/\sqrt{c}$ & $n\eps^2\ge 2^{12}\log\frac{16}{\delta}$\\[2pt]
Thm.~\ref{thm:logconcave} & $c$-exponential concentration & $L_0/\sqrt{c}$ & $n\eps^2\ge 2^{12}\log\frac{16}{\delta}$,\newline $n\ge Q/61$\\[2pt]
Prop.~\ref{prop:lcbeyond} & $c$-exponential concentration & $\frac{8\sqrt{nQ}}{n+Q}\,L_0/\sqrt{c}$ & $n\eps^2\ge 2^{12}\log\frac{16}{\delta}$\\[2pt]
Cor.~\ref{rem:lcpsialpha} & $\psi_\alpha$ concentration, $\alpha\in[1,2]$ & $L_0/\sqrt{c}$ & $n\eps^2\ge 2^{12}\log\frac{16}{\delta}$,\newline $n\ge Q/61$\\[2pt]
Cor.~\ref{cor:logconcave} (i) & log-concave, $\norm{\mathrm{Cov}}_{\mathrm{op}}\le\kappa/d$ & $L_0/\sqrt{C_0\kappa\log d}$ & $n\eps^2\ge 2^{12}\log\frac{16}{\delta}$,\newline $n\ge Q/61$\\[2pt]
Cor.~\ref{cor:logconcave} (ii) & as (i), granting \citet{BKL26} & $L_0/\sqrt{9C_{\mathrm{BKL}}\kappa}$ & $n\eps^2\ge 2^{12}\log\frac{16}{\delta}$,\newline $n\ge Q/61$\\
\bottomrule
\end{tabularx}
}
\caption{The laws of robustness of this paper; $Q$ and the shared rate factor $L_0$ are
defined in \eqref{eq:Q}. Here $c$ is the isoperimetric (respectively exponential- or
$\psi_\alpha$-concentration) constant, and $C_0$ and $C_{\mathrm{BKL}}$ are the absolute
constants of Corollary~\ref{cor:logconcave}. The last row, corresponding to Corollary~\ref{cor:logconcave} (ii), rests on an announced, not yet
refereed, theorem (Remark~\ref{rem:kls}).}
\label{tab:results}
\end{table}

\paragraph{Scope and finite-sample conditions.} Except for the Euclidean log-concave
specialization (Corollary~\ref{cor:logconcave}), the arguments use only metric
concentration (Remark~\ref{rem:metric}). The cap $Q\le61\,n$ is a capacity burn-in: it
makes the net-indexed tail bound beat the net entropy without changing the Lipschitz rate.

\paragraph{Companion paper.} Theorem~\ref{thm:master} is stated in the generality of a
companion paper \citep{PeterModular}, which uses the same event algebra for three further
relaxations. For heterogeneous mixtures, the weighted average of the component
isoperimetric constants replaces the worst one. For vector-valued predictors, the bound at
the Frobenius Lipschitz scale is free of the output dimension. For stationary Markov
covariates with labels conditionally independent given the trajectory, the i.i.d.\
threshold survives. The companion also extends this result to stationary
$\beta$-mixing covariates via Berbee coupling and blocking, with the same threshold
under additional sample-size and coupling conditions. The present paper uses only the
scalar i.i.d.\ case of that theorem and is self-contained.

\subsection{Related Work}\label{sec:related}
A law of robustness was conjectured by Bubeck, Li and Nagaraj \citeyearpar{BLN21} for
two-layer networks with arbitrary weights and Gaussian data;
\citet[Corollary~1.3]{Shmalo26}, in a July 2026 preprint, proves the
conjectured law up to one logarithmic factor for every continuous piecewise-linear
activation. The universal form \eqref{eq:BSlaw}, our starting point, is due to Bubeck and
Sellke (\citeyear{BS21}; journal version \citeyear{BS23}), who also noted that under only a
Poincar\'e inequality their argument yields a weaker law via exponential concentration
(\citealp{BS21}, end of \S1.2); the content of Section~\ref{sec:logconcave} is that the
\emph{unweakened} rate $\sqrt{nd/Q}$ survives, at the price of the burn-in. Wu, Huang and
Zhang obtained the distribution-free $\Omega(\eps n^{1/d})$ bound on bounded supports
\citeyearpar[Theorem~3.9]{WHZ23} and, for parametrizations with $WJ$ polynomially bounded, an
$\widetilde\Omega(\eps\sqrt{n/p})$ law on arbitrary bounded supports \citeyearpar[Theorem~3.4]{WHZ23};
the $d$ of \eqref{eq:BSlaw} is the price of dropping isoperimetry. On a bounded support
neither bound dominates the other uniformly: the packing bound is parameter-free but grows
only like $n^{1/d}$, whereas ours grows like $\sqrt n$ and weakens as $Q$ grows. They also show \citep[Theorem~D.4 of arXiv:2202.11592v1]{WHZ23} that fitting below the
noise level at all requires $p=\widetilde\Omega(n\eps^2)$ parameters, an observation for
which we prove a self-contained quantitative version (Proposition~\ref{prop:overparam}).

On the technical side, recent progress on the KLS conjecture feeds directly into the
log-concave extension: the best published Poincar\'e bound for isotropic log-concave
measures \citep[Theorem~1.2 and Eq.~(1.4)]{Klartag23} yields the logarithmic dimension
dependence in Corollary~\ref{cor:logconcave}(i). \citet[Theorem~1.1]{BKL26} have since
announced a proof of the full conjecture, $C_P(\mu)\le C\norm{\mathrm{Cov}(\mu)}_{\mathrm{op}}$
for every log-concave $\mu$, whose crucial step is a criterion of \citet{SongZhang26}, who
also claim the full conjecture. This gives the dimension-free part~(ii) of the corollary.

Architecture-specific work shows that the universal lower bound need not be sufficient for
robustness: in their high-dimensional empirical-risk-minimization models and for their
sensitivity metric, random features remain non-robust at every overparametrization level,
while even-activation NTK models meet the law once $p\gtrsim nd$ \citep{BKM23}. Kumar, Luo
and Sellke \citeyearpar{KLS24} extended the law to sparse networks using an effective
parameter count that includes the mutual information between the pruning mask and the
data.

\section{Preliminaries}\label{sec:prelim}

This section introduces the notation, concentration assumptions, and probabilistic tools used throughout the paper.

\subsection{Basic Notation}
Let $\norm{\cdot}$ denote the Euclidean norm (on $\R^d$ or $\R^K$, per context) and
$\mathbb{B}_2^K=\{v\in\R^K:\norm v\le1\}$ the unit ball in $\R^K$. For $f:\R^d\to\R^K$,
$\Lip(f)\coloneqq\sup_{x\ne x'}\norm{f(x)-f(x')}/\norm{x-x'}$, and for a vector-valued function 
$f=(f_1,\dots,f_K)$ we write $\LF(f)\coloneqq(\sum_{j=1}^K\Lip(f_j)^2)^{1/2}$; note
\begin{equation}\label{eq:frobop}
 \Lip(f)\;\le\;\LF(f)\;\le\;\sqrt K\,\Lip(f).
\end{equation}
Indeed, for any $x,x'$ we have
$\norm{f(x)-f(x')}^2=\sum_j(f_j(x)-f_j(x'))^2\le \sum_j \Lip(f_j)^2\norm{x-x'}^2
=\LF(f)^2\norm{x-x'}^2$, so $\Lip(f)\le\LF(f)$. Conversely $\Lip(f_j)\le\Lip(f)$ for each $j$
(coordinate projection is $1$-Lipschitz), whence $\LF(f)\le\sqrt K\Lip(f)$.

Two bookkeeping quantities recur in every statement below and are used globally: the
\emph{complexity term} $Q=p\log(48WJ/\eps)+\log(16/\delta)$ of \eqref{eq:Q}, and the
\emph{variance proxy} $v=c\Lip^2/d$ attached to a $c$-isoperimetric measure
(Definition~\ref{def:isop} below), the scale at which every pairing bound of
Section~\ref{sec:skeleton} is calibrated.

\subsection{Isoperimetry and Parametrized Classes}

\begin{definition}[$c$-isoperimetry, \citealp{BS21}]\label{def:isop}
A probability measure $\mu$ on $\R^d$ satisfies \emph{$c$-isoperimetry} if for every bounded
$L$-Lipschitz $f:\R^d\to\R$ and every $t\ge0$,
\begin{equation}\label{eq:isop}
 \Prob_{X\sim\mu}\big[|f(X)-\E_\mu f|\ge t\big]\;\le\;2\exp\!\Big(-\frac{d\,t^2}{2cL^2}\Big).
\end{equation}
We read the boundary case $c=0$ as the degenerate limit in which the right-hand side of
\eqref{eq:isop} is $0$ for every $t>0$; equivalently, $f=\E_\mu f$ holds $\mu$-a.s.\ for every
bounded Lipschitz $f$, i.e.\ $\mu$ is a Dirac mass. We denote by $v\coloneqq cL^2/d$ the 
\emph{variance proxy} attached to a $c$-isoperimetric measure, so that \eqref{eq:isop} reads
\[
 \Prob_{X\sim\mu}\big[|f(X)-\E_\mu f|\ge t\big]\;\le\;2\exp\!\Big(-\frac{t^2}{2v}\Big).
\]
\end{definition}

Note that $O(1)$-isoperimetry is not a narrow hypothesis. The canonical example is
the isotropic Gaussian $N(0,\tfrac1d \Id_d)$, via the Gaussian concentration inequality
\citep{Ledoux01,Ver26}. We record one closure property, elementary from
Definition~\ref{def:isop}: an $L_0$-Lipschitz pushforward of an $O(1)$-isoperimetric
measure is again $O(1)$-isoperimetric, with the constant rescaled by $L_0^2$. Further
examples (the uniform laws on the sphere and on the solid cube, uniformly convex
potentials, and positively curved compact Riemannian manifolds) are referenced in
Appendix~\ref{app:toolbox}.

\begin{remark}[Metric probability spaces]\label{rem:metric}
Definitions~\ref{def:isop} and~\ref{def:expconc} use only a metric. Hence
Theorems~\ref{thm:baseline} and~\ref{thm:logconcave}, which use the domain only through these
concentration hypotheses, extend to a metric probability space $(M,d_M,\mu)$ (the setting
of \citealp[Definition~3.3]{Peter}), with $\Lip$ taken with respect to $d_M$ and the
dimension factor treated as an effective parameter $d_{\mathrm{eff}}$. Peter's
concentration parameter corresponds to $c/d_{\mathrm{eff}}$ in the normalization of
Definition~\ref{def:isop}. Euclidean consequences such as Corollary~\ref{cor:logconcave}
do not transfer.
\end{remark}

We use an uncentered version of the tail classes of \citet{Peter}.

\begin{definition}[$\alpha$-sub-exponential tail classes; cf.\ {\citealp[Definition~2.3]{Peter}}]
\label{def:exptail}
For $\alpha,s>0$ and $R\ge1$, a real random variable $A$ belongs to the class
$\expTail^{\alpha}(R,s)$ if
\begin{equation}\label{eq:exptail}
 \Prob\big[|A|\ge t\big]\;\le\;2R\,\exp\!\big(-(t/s)^{\alpha}\big)
 \qquad\text{for all }t\ge0.
\end{equation}
We call \eqref{eq:exptail} a \emph{$\psi_\alpha$ tail} with scale $s$, following the usual
Orlicz terminology; $\alpha=2$ is sub-Gaussian and $\alpha=1$
sub-exponential. The two concentration hypotheses in force below are stated at the
endpoints of the scale---$\alpha=2$ in Definition~\ref{def:isop}, $\alpha=1$ in
Definition~\ref{def:expconc}---and the intermediate range $\alpha\in(1,2)$ enters only
through Lemmas~\ref{lem:psimom} and~\ref{lem:keycap}, which extend
Section~\ref{sec:logconcave} to the whole range $\alpha\in[1,2]$ at no cost
(Corollary~\ref{rem:lcpsialpha}).
\end{definition}

In this notation, Definition~\ref{def:isop} states that
$f(x)-\E_\mu f\in\expTail^{2}(1,\sqrt{2v})$
with $v=cL^2/d$, for every bounded $L$-Lipschitz $f$; the exponential-concentration
hypothesis of Section~\ref{sec:logconcave} (Definition~\ref{def:expconc}) states that
$f(x)-\E_\mu f\in\expTail^{1}(1,\sqrt v)$.

\begin{definition}[$(p,J,W)$-parametrization]\label{def:param}
A class $\cF$ of functions $\R^d\to\R^K$ ($K\ge1$) admits a \emph{$(p,J,W)$-parametrization} if
$\cF=\{f_w : w\in\cW\}$ for some $\cW\subseteq\R^p$ with $\operatorname{diam}(\cW)\le W$ in some
norm $|\cdot|$ on $\R^p$, and
\[
 \norm{f_w-f_{w'}}_\infty\;\coloneqq\;\sup_{x\in\R^d}\,\norm{f_w(x)-f_{w'}(x)}_2\;\le\;J\,|w-w'|
 \qquad\text{for all }w,w'\in\cW .
\]
We assume throughout $W,J\ge1$.
\end{definition}

\subsection{Probabilistic Toolbox}\label{sec:toolbox}

We first recall some classical inequalities and then state a few elementary lemmas. 
The proofs of the lemmas are deferred to Appendix~\ref{app:toolbox}.

\begin{fact}[Hoeffding's lemma; {\citealp[Lemma~2.2]{BLM13}}]\label{fact:hoeffding}
If $V$ is a real random variable with $\E V=0$ and $V\in[a,b]$ almost surely, then
$\E e^{\lambda V}\le e^{\lambda^2(b-a)^2/8}$ for all $\lambda\in\R$.
\end{fact}

\begin{fact}[One-sided Bernstein inequality {\citealp[Thm.~2.9 and Eq.~(2.10)]{BLM13}}]\label{fact:bernstein}
Let $U_1,\dots,U_n$ be independent with $\E U_i=0$, $U_i\le B$ a.s.\ and
$\sum_i\E U_i^2\le n\bar u$. Then for $t\ge0$,
\begin{equation}
 \Prob\!\left[\sum_i U_i\ge t\right] \le \exp\left(-\frac{t^2}{2(n\bar u+Bt/3)}\right).
\end{equation}
\end{fact}

\begin{lemma}[Conditionally centered bounded sums]\label{lem:azuma}
Let $(\mathcal G_i)_{i=0}^n$ be a filtration and let $V_1,\dots,V_n$ be random variables such
that $V_i$ is $\mathcal G_i$-measurable, $\E[V_i\,|\,\mathcal G_{i-1}]=0$ and $|V_i|\le B$
almost surely, where $B\ge0$. Then for every $s\ge0$, it holds that
$\Prob[\sum_{i=1}^n V_i\ge s]\le\exp\left(-\frac{s^2}{2nB^2}\right)$, where for $B=0$ the
right-hand side is read as $1$ if $s=0$ and as $0$ if $s>0$.
\end{lemma}
\begin{proof}
See \hyperref[proof:azuma]{Appendix~\ref*{app:toolbox}}.
\end{proof}

Section~\ref{sec:logconcave} yields an MGF only on a bounded interval. The next lemma
turns such a capped MGF into the quadratic and linear tail branches used there.

\begin{lemma}[Two-branch Chernoff bound under a capped MGF]\label{lem:cappedtail}
Let $W$ be a real random variable, and let $a>0$ and $\theta_0\in(0,\infty)$ be such that
\[
 \E\big[e^{\theta W}\big]\;\le\;e^{a\theta^2}
 \qquad\text{for all }0\le\theta\le\theta_0 .
\]
Then for every $s\ge0$,
\[
 \Prob[W\ge s]\;\le\;
 \begin{cases}
 \exp\!\Big(-\dfrac{s^2}{4a}\Big), & \text{if } s\le 2a\theta_0,\\[8pt]
 \exp\!\big(a\theta_0^2-\theta_0 s\big), & \text{always.}
 \end{cases}
\]
\end{lemma}
\begin{proof}
See \hyperref[proof:cappedtail]{Appendix~\ref*{app:toolbox}}.
\end{proof}

The pairing arguments need the second and fourth moments of a sub-Gaussian fluctuation
and two exponential moments of its square. We record the three conversions together.

\begin{lemma}[Moments and exponential moments from sub-Gaussian tails]\label{lem:subgsq}
Consider a random variable $A\in\expTail^{2}(1,\sqrt{2v})$ for some $v>0$, i.e.\
$\Prob[|A|\ge t]\le 2e^{-t^2/(2v)}$ for all $t\ge0$. Then
\begin{enumerate}[label=(\alph*)]
 \item $\E\,e^{A^2/(4v)}\le 3$,
 \item $\E A^{2k}\le 2\,k!\,(2v)^k$ for all $k\ge1$,
 \item $\E\,e^{\theta A^2}\le 1+\frac{4\theta v}{1-2\theta v}$ for $0\le\theta<\tfrac1{2v}$.
\end{enumerate}
\end{lemma}
\begin{proof}
See \hyperref[proof:subgsq]{Appendix~\ref*{app:toolbox}}.
\end{proof}

The next lemma is the $\psi_\alpha$ counterpart of Lemma~\ref{lem:subgsq}~(b) across the
whole range $\alpha\in[1,2]$.

\begin{lemma}[Moments from $\psi_\alpha$ tails]\label{lem:psimom}
Let $\alpha\in[1,2]$ and let $B\in\expTail^{\alpha}(1,s)$ for some $s>0$, i.e.\
$\Prob[|B|\ge t]\le2e^{-(t/s)^{\alpha}}$ for all $t\ge0$. Then for every $k\ge1$,
\[
 \E|B|^{k}\;\le\;2\,s^{k}\,\Gamma\!\Big(\frac k\alpha+1\Big)\;\le\;2\,s^{k}\,(k!)^{1/\alpha}.
\]
\end{lemma}
\begin{proof}
See \hyperref[proof:psimom]{Appendix~\ref*{app:toolbox}}.
\end{proof}

\section{The Modular Skeleton and a Quantitative Baseline}\label{sec:skeleton}

This section develops the product-MGF estimates and modular master theorem used in
both robustness laws. The truncation reduction, scalar noise-correlation
decomposition, and parameter-net argument are adapted from the proofs of Lemma~2.1
and Theorems~2--3 of \citet{BS21}. We give explicit product-MGF formulations, including
a capped form for $\psi_\alpha$ fluctuations, and state the common reduction through
separate probabilistic conditions. These formulations allow the exponential-concentration
argument of Section~\ref{sec:logconcave}. The baseline theorem recovers the
Bubeck--Sellke scalar, single-component law with adjusted constants.

\subsection{The Regression Model}

We first fix the scalar i.i.d.\ model that supplies the baseline and the common residual
notation for Sections~\ref{sec:skeleton} and~\ref{sec:logconcave}.

\begin{assumption}[Scalar i.i.d.\ model]\label{ass:model}
Suppose that $(x_i,y_i)_{i=1}^n$ are i.i.d.\ pairs in $\R^d\times[-1,1]$ with $x_i\sim\mu$. Write
$g(x)\coloneqq\E[y\,|\,x]$, and $z_i\coloneqq y_i-g(x_i)\in[-2,2]$ for the corresponding residuals. 
The expected noise is assumed to be non-zero:
\[
 \sigma^2\coloneqq\E\big[\Var[y\,|\,x]\big]=\E\,z_i^2\;\in(0,1].
\]
We fix an overfitting margin $\eps\in(0,\sigma^2]$ and a failure probability $\delta\in(0,1)$.
\end{assumption}

We will repeatedly use the truncation $\tau:\R\to[-1,1]$, $\tau(u)=\max(-1,\min(1,u))$.

\begin{lemma}[Truncation transfer]\label{lem:trunc}
For all $u,u'\in\R$, $y\in[-1,1]$ and $f,f':\R^d\to\R$:
\[
 |\tau(u)-y|\;\le\;|u-y|,\qquad
 \Lip(\tau\circ f)\;\le\;\Lip(f),\qquad
 \norm{\tau\circ f-\tau\circ f'}_\infty\;\le\;\norm{f-f'}_\infty .
\]
\end{lemma}
\begin{proof}
The truncation $\tau$ is $1$-Lipschitz
(it is the composition of $u\mapsto\min(1,u)$ and $u\mapsto\max(-1,u)$, each
$1$-Lipschitz) and it fixes every point of $[-1,1]$. Hence
$|\tau(u)-y|=|\tau(u)-\tau(y)|\le|u-y|$ for $y\in[-1,1]$, and the second and third
inequalities follow by composing with $f$, $f'$.
\end{proof}

Since $|y_i|\le1$, the first inequality shows that truncation can only decrease the
training error. Consequently it suffices to prove all lower bounds for $[-1,1]$-valued
functions; the bounds transfer verbatim to the original class because a fit by $f$ implies
a fit by $\tau\circ f$ and $\Lip(f)\ge\Lip(\tau\circ f)$.

\subsection{The Lemmas}

The first two lemmas control function-independent quantities.

\begin{lemma}[Empirical noise level]\label{lem:noise}
Under Assumption~\ref{ass:model}, for every $\xi\in(0,1]$,
\[
 \Prob\Big[\frac1n\sum_{i=1}^n z_i^2<\sigma^2-\frac{\xi}{4}\Big]
 \;\le\;\exp\!\Big(-\frac{n\xi^2}{144}\Big).
\]
\end{lemma}
\begin{proof}
Apply Fact~\ref{fact:bernstein} to $U_i=\sigma^2-z_i^2$, which satisfies $\E U_i=0$,
$U_i\le\sigma^2\le1 \eqqcolon B$ and $\E U_i^2\le\E z_i^4\le 4\E z_i^2\le4 \eqqcolon \bar u$. With $t=n\xi/4$, it follows that
\[
 \Prob\Big[\sum_iU_i\ge \frac{n\xi}{4}\Big]
 \le \exp\!\Big(-\frac{(n\xi/4)^2}{2(4n+\frac{n\xi}{12})}\Big)
 \le \exp\!\Big(-\frac{n\xi^2}{32\,(4+\tfrac1{12})}\Big)
 \le \exp\!\Big(-\frac{n\xi^2}{144}\Big). \qedhere
\]
\end{proof}

The two remaining function-independent sums are controlled in the following
distribution-free filtration form, which is the form in which they are used in every
theorem below (with the conditioning $\sigma$-algebra $\cG_0$ generated by the covariates---for dependent data, by the whole trajectory).

\begin{lemma}[Global noise sums]\label{lem:globalsums}
Let $K\ge1$, let $\cG_0$ be a $\sigma$-algebra, let $g:\R^d\to \mathbb{B}_2^K$ be measurable with
$(x_i)_{i\le n}$ $\cG_0$-measurable, and let $z_1,\dots,z_n\in2\mathbb{B}_2^K$ be random vectors
such that, with $\cG_i\coloneqq\sigma(\cG_0,z_1,\dots,z_i)$,
$\E[z_i\,|\,\cG_{i-1}]=0$ for each $i\le n$. Then for every $\eps>0$:
\begin{enumerate}[label=\emph{(\roman*)},leftmargin=2em,itemsep=1pt]
\item if $K=1$,
$\ \Prob\big[\,\big|\sum_{i}z_i\big|>\frac{n\eps}{16}\big]\le2e^{-n\eps^2/2048}$;
\item $\Prob\big[\sum_{i}\inner{z_i}{-g(x_i)}>\frac{n\eps}{16}\big]\le e^{-n\eps^2/2048}$.
\end{enumerate}
\end{lemma}
\begin{proof}
\emph{(i)} Lemma~\ref{lem:azuma} with $V_i=z_i$ and with $V_i=-z_i$ ($B=2$,
$s=\frac{n\eps}{16}$, so $\frac{s^2}{2nB^2}=\frac{n\eps^2}{2048}$), and a union bound over
the two signs. \emph{(ii)} Lemma~\ref{lem:azuma} with
$V_i\coloneqq\inner{z_i}{-g(x_i)}$, which is $\cG_i$-measurable with
$|V_i|\le\norm{z_i}\norm{g(x_i)}\le2$ and
$\E[V_i\,|\,\cG_{i-1}]=\inner{\E[z_i\,|\,\cG_{i-1}]}{-g(x_i)}=0$, since $g(x_i)$ is
$\cG_0$-measurable.
\end{proof}

The preceding lemmas control sample-only events. The next lemma is a deterministic
vector formulation of the loss expansion in the proof of
\citet[Lemma~2.1]{BS21}: on every sample,
a fit below the noise level forces the pairing $\sum_i\inner{z_i}{f(x_i)-g(x_i)}$ above $\tfrac38 n\xi$. 
The master theorem (Theorem~\ref{thm:master}) bounds the same sum from above, with high probability 
and uniformly over a finite net. The two bounds are incompatible, so a below-noise fit will force a large 
Lipschitz constant.

\begin{lemma}[Overfitting decomposition]\label{lem:decomp}
Let $K\ge1$, let $f,g:\R^d\to\R^K$ be measurable, let
$(x_i,y_i)_{i\le n}\subset\R^d\times\R^K$, and set $z_i\coloneqq y_i-g(x_i)$. For all
$\sigma^2\ge0$ and $\xi>0$: if $\frac1n\sum_i\norm{z_i}^2\ge\sigma^2-\xi/4$ and
$\frac1n\sum_i \norm{f(x_i)-y_i}^2\le\sigma^2-\xi$, then
\[
 \sum_{i=1}^n \inner{z_i}{\,f(x_i)-g(x_i)}\;\ge\;\frac{3}{8}\,n\xi .
\]
\end{lemma}
\begin{proof}
Write $h\coloneqq f-g$. Since $f(x_i)-y_i=h(x_i)-z_i$,
bilinearity of the inner product on $\R^K$ gives, for each $i$,
\[
 \norm{f(x_i)-y_i}^2=\norm{h(x_i)-z_i}^2
 =\norm{h(x_i)}^2-2\inner{z_i}{h(x_i)}+\norm{z_i}^2 .
\]
Summing over $i$ and rearranging,
\begin{align*}
  2\sum_{i=1}^n\inner{z_i}{h(x_i)}
  &=\sum_{i=1}^n\norm{h(x_i)}^2+\sum_{i=1}^n\norm{z_i}^2-\sum_{i=1}^n\norm{f(x_i)-y_i}^2 \\
  &\ge\;0+n\Big(\sigma^2-\frac\xi4\Big)-n\big(\sigma^2-\xi\big)\;=\;\frac34\,n\xi\,.
\end{align*}
Dividing by $2$ gives the claim; no boundedness of $f$, $y_i$ or $z_i$
is used.
\end{proof}

The key probabilistic step bounds a centered noise--fluctuation product. We first state a
vector template with an all-$\lambda$ MGF, then specialize it to scalar sub-Gaussian
fluctuations.

\begin{lemma}[Product MGF from square-exponential moments; unconditional form]
\label{lem:keymaster}
Let $K\ge1$ and let $A$ be an $\R^K$-valued random vector such that, for some $v>0$,
$M\ge1$ and constants $m_1>0$, $m_2\in(0,4]$,
\begin{equation}\label{eq:sqmom}
 \E\,e^{\norm A^2/(4v)}\;\le\;M
 \qquad\text{and}\qquad
 \E\,e^{\theta\norm A^2}\;\le\;1+\frac{m_1\,\theta v}{1-m_2\,\theta v}
 \quad\text{for all }0\le\theta<\frac1{m_2v}.
\end{equation}
Let $Z$ be an $\R^K$-valued random variable with $\norm Z\le2$ a.s.\ and $\E[Z\,|\,A]=0$.
Then for every $\lambda\in\R$,
\[
 \E\,e^{\lambda\inner ZA}\;\le\;\exp\big(C\,\lambda^2v\big),
 \qquad
 C\;\coloneqq\;\max\Big\{\frac{2m_1}{1-m_2/8},\;4+16\log M\Big\}.
\]
\end{lemma}
\begin{proof}
By Cauchy--Schwarz, $|\inner ZA|\le\norm Z\norm A\le2\norm A$, so conditionally on $A$ the
variable $\lambda\inner ZA$ is centered with range at most $4|\lambda|\norm A$; by
Fact~\ref{fact:hoeffding} applied conditionally,
\begin{equation}\label{eq:condhoeffmgf}
 \E\big[e^{\lambda\inner ZA}\,\big|\,A\big]\;\le\;e^{2\lambda^2\norm A^2}.
\end{equation}
\emph{Case 1: $\lambda^2\le\frac1{16v}$.} Then $\theta\coloneqq2\lambda^2\le\frac1{8v}$,
so $m_2\theta v\le m_2/8\le\frac12<1$ and the second bound in \eqref{eq:sqmom} applies:
\[
 \E\,e^{2\lambda^2\norm A^2}\;\le\;1+\frac{m_1\theta v}{1-m_2/8}
 \;=\;1+\frac{2m_1\,\lambda^2v}{1-m_2/8}
 \;\le\;\exp\Big(\frac{2m_1}{1-m_2/8}\,\lambda^2v\Big).
\]
\emph{Case 2: $\lambda^2>\frac1{16v}$.} By the inequality of arithmetic and geometric
means, $\lambda\inner ZA\le2|\lambda|\norm A\le4\lambda^2v+\frac{\norm A^2}{4v}$, hence by
the first bound in \eqref{eq:sqmom},
\[
 \E\,e^{\lambda\inner ZA}\;\le\;e^{4\lambda^2v}\,\E\,e^{\norm A^2/(4v)}
 \;\le\;M\,e^{4\lambda^2v}\;=\;e^{4\lambda^2v+\log M}\;\le\;e^{(4+16\log M)\lambda^2v},
\]
using $16\lambda^2v>1$. In both cases, the expectation is at most $e^{C\lambda^2v}$.
\end{proof}

Applying Lemma~\ref{lem:keymaster} with the moment bounds of Lemma~\ref{lem:subgsq}
gives the scalar product MGF used in the baseline law. The centered scalar product
is also controlled in the proof of \citet[Theorem~2]{BS21}, using sub-Gaussian tails
and stability under independent summation. Here we state the bound explicitly in MGF form.
Under weaker concentration, we use its capped counterpart,
given in Lemma~\ref{lem:keycap}.

\begin{lemma}[Product MGF; key lemma]\label{lem:key}
Let $A\in\expTail^{2}(1,\sqrt{2v})$ (Definition~\ref{def:exptail}) for
some $v>0$, and let $Z$ be a real random variable with $|Z|\le2$ a.s.\ and
$\E[Z\,|\,A]=0$. Then for every $\lambda\in\R$,
\[
 \E\,e^{\lambda ZA}\;\le\;\exp\!\big(22\,\lambda^2v\big).
\]
(For $v=0$, i.e.\ $A=0$ a.s., the bound holds trivially.)
\end{lemma}
\begin{proof}
Apply Lemma~\ref{lem:keymaster} with $K=1$, so that $\norm A^2=A^2$:
Lemma~\ref{lem:subgsq}(a) supplies the first hypothesis of \eqref{eq:sqmom} with $M=3$, and
Lemma~\ref{lem:subgsq}(c) the second with $m_1=4$, $m_2=2$. The resulting constant is
\begin{equation*}
  C=\max\{\frac{2\cdot4}{1-1/4},\,4+16\log3\}=\max\{\frac{32}3,\,4+16\log3\} =4+16\log3\le22\,.\qedhere
\end{equation*}
\end{proof}

The capped variant below holds only on a bounded range of $\lambda$, and it is the form
available when the covariate fluctuation has a $\psi_\alpha$ tail with $\alpha<2$. 

\begin{lemma}[{Product MGF; capped form, $\alpha\in[1,2]$}]\label{lem:keycap}
Let $K\ge1$, let $\alpha\in[1,2]$, and set $\eta_\alpha\coloneqq1-\frac1\alpha\in[0,\frac12]$.
Let $A$ be an $\R^K$-valued random vector obeying, for some $s>0$, the moment bound
\begin{equation}\label{eq:psimoments}
 \E\norm A^{k}\;\le\;2\,s^{k}\,(k!)^{1/\alpha}\qquad\text{for every }k\ge1,
\end{equation}
for which $\norm A\in\expTail^{\alpha}(1,s)$ (Definition~\ref{def:exptail}) is sufficient by
Lemma~\ref{lem:psimom}; and let $Z$ be $\R^K$-valued with
$\norm Z\le2$ a.s.\ and $\E[Z\,|\,A]=0$. Then
\[
 \E\,e^{\lambda\inner ZA}\;\le\;\exp\big(2^{4-\eta_\alpha}\,\lambda^2s^2\big)
 \qquad\text{for all }|\lambda|\;\le\;\frac{3^{\eta_\alpha}}{4s}.
\]
(For $s=0$, i.e.\ $A=0$ a.s., the bound holds trivially.)
\end{lemma}
\begin{proof}
Put $q\coloneqq2|\lambda|s\le3^{\eta_\alpha}/2\le\sqrt3/2<1$ and $a_k\coloneqq q^k(k!)^{-\eta_\alpha}$.
By Cauchy--Schwarz we have $|\inner ZA|\le2\norm A$, so \eqref{eq:psimoments} gives
$\E|\inner ZA|^k\le2^k\,\E\norm A^k\le2^{k+1}s^k(k!)^{1/\alpha}$ for every $k\ge1$, whence
\[
 \E\,e^{|\lambda\inner ZA|}\;\le\;1+2\sum_{k\ge1}a_k\;\le\;1+2\sum_{k\ge1}q^k\;<\;\infty
\]
because $(k!)^{-\eta_\alpha}\le1$ and $q<1$; the exponential series may therefore be integrated
term by term. The linear term vanishes,
$\E[\lambda\inner ZA]=\lambda\,\E\big[\inner{\E[Z\,|\,A]}{A}\big]=0$, so
\[
 \E\,e^{\lambda\inner ZA}\;\le\;1+\sum_{k\ge2}\frac{|\lambda|^k\,\E|\inner ZA|^k}{k!}
 \;\le\;1+2\sum_{k\ge2}a_k .
\]
For $k\ge2$, the ratio $a_{k+1}/a_k=q\,(k+1)^{-\eta_\alpha}$ is decreasing in $k$, hence at
most $q\,3^{-\eta_\alpha}\le\frac12$; therefore
$\sum_{k\ge2}a_k\le a_2\sum_{j\ge0}2^{-j}=2\,a_2=2^{1-\eta_\alpha}q^2$, and
\[
 \E\,e^{\lambda\inner ZA}\;\le\;1+2^{2-\eta_\alpha}q^2\;=\;1+2^{4-\eta_\alpha}\lambda^2s^2
 \;\le\;e^{2^{4-\eta_\alpha}\lambda^2s^2}. \qedhere
\]
\end{proof}

\begin{corollary}[Pairing tail bound]\label{cor:pairing}
In the setting of Assumption~\ref{ass:model}, let $\alpha:\R^d\to\R$ be measurable with
$\alpha(X)\in\expTail^{2}(1,\sqrt{2v})$ for $X\sim\mu$, and set $A_i=\alpha(X_i)$ for $X_i\sim\mu$ i.i.d. Then for every
$s\ge0$,
\[
 \Prob\Big[\sum_{i=1}^nz_iA_i\ge s\Big]\;\le\;\exp\!\Big(-\frac{s^2}{88\,n\,v}\Big).
\]
\end{corollary}
\begin{proof}
The pairs $(z_i,A_i)$ are i.i.d., $|z_i|\le2$ and $\E[z_i\,|\,A_i]=\E[\E[z_i|x_i]\,|\,A_i]=0$,
so by independence across $i$ and Lemma~\ref{lem:key},
$\E e^{\lambda\sum_iz_iA_i}\le e^{22n\lambda^2v}$. Markov's inequality with
$\lambda=s/(44nv)$ gives the claim.
\end{proof}

Uniformity over $\cF$ is obtained by discretizing the predictors below the candidate
Lipschitz threshold, following the proof of \citet[Theorem~3]{BS21}.
The net preserves a below-noise fit at cost $\eps/2$ and has entropy
at most $p\log(48WJ/\eps)$.

\begin{lemma}[Discretization]\label{lem:net}
Let $\cF=\{f_w:w\in\cW\}$ admit a $(p,J,W)$-parametrization (Definition~\ref{def:param},
scalar or vector valued), let $\tau$ denote the relevant truncation (clipping to $[-1,1]$ in the
scalar case; the identity in the vector case treated in \citet{PeterModular}, where $\cF$ is assumed
$\mathbb{B}_2^K$-valued). Let
$\mathrm{Lip}_\bullet\in\{\Lip,\LF\}$ be the Lipschitz functional under consideration,
$L\in(0,\infty]$, $\eps\in(0,1]$, and $S_L\coloneqq\{w\in\cW:\mathrm{Lip}_\bullet(\tau f_w)\le L\}$,
so that $S_\infty=\cW$. There exists a \emph{parameter net}
$\cN_L^{\cW}\subseteq S_L$ with
\[
 |\cN_L^{\cW}|\;\le\;\Big(\frac{48\,WJ}{\eps}\Big)^{p}
\]
such that for every $w\in S_L$ there is $w'\in\cN_L^{\cW}$ with
$\norm{\tau f_w-\tau f_{w'}}_\infty\le\eps/16$. Write
\[
 \cN_L\;\coloneqq\;\{\tau f_{u}:u\in\cN_L^{\cW}\}
\]
for the induced \emph{truncated function net}: every $f'\in\cN_L$ satisfies
$\mathrm{Lip}_\bullet(f')\le L$, and $|\cN_L|\le|\cN_L^{\cW}|\le(48WJ/\eps)^p$.
Moreover, on any sample $(x_i,y_i)_{i\le n}$ with $\norm{y_i}\le1$ for all $i$: if some
$w\in S_L$ satisfies
$\frac1n\sum_i\norm{\tau f_w(x_i)-y_i}^2\le\sigma^2-\eps$, then $w'\in\cN_L^{\cW}$ as
above may be chosen so that $f'\coloneqq\tau f_{w'}\in\cN_L$ satisfies
$\frac1n\sum_i\norm{f'(x_i)-y_i}^2\le\sigma^2-\eps/2$.
\end{lemma}
\begin{proof}
Let $\eta\coloneqq\eps/(16J)$. A subset of $S_L$ is \emph{$\eta$-separated} if any two of its
distinct points are at distance $>\eta$ in the parameter norm $|\cdot|$
\citep[Definition~4.2.4]{Ver26}.
For the cardinality: the closed balls of radius $\eta/2$
around the points of an $\eta$-separated subset of $S_L$ are disjoint
\citep[Remark~4.2.5]{Ver26} and contained in the ball of radius $W+\eta/2$ about any
fixed point of $\cW$, since $W\geq\operatorname{diam}(\cW)$. Comparing volumes in
$(\R^p,|\cdot|)$ for each finite subfamily, every
$\eta$-separated subset of $S_L$ has at most
$N\coloneqq\big(\frac{W+\eta/2}{\eta/2}\big)^p\le\big(\frac{3W}{\eta}\big)^p
=\big(\frac{48WJ}{\eps}\big)^p$ points, using $\eta\le W$ since $W,J\geq 1$.
Hence an $\eta$-separated subset $\cN_L^{\cW}\subseteq S_L$ of maximum cardinality exists
($\emptyset$ is $\eta$-separated, and cardinalities are bounded by $N$). It follows 
that every $w\in S_L$ has $|w-w'|\le\eta$ for some $w'\in\cN_L^{\cW}$. Since $\tau$ is
$1$-Lipschitz in sup norm, it follows that
\[
 \norm{\tau f_w-\tau f_{w'}}_\infty\;\le\;\norm{f_w-f_{w'}}_\infty\;\le\;J|w-w'|\;\le\;J\eta\;=\;\frac{\eps}{16}.
\]
The map $u\mapsto\tau f_{u}$ sends $\cN_L^{\cW}$ onto $\cN_L$, so $|\cN_L|\le|\cN_L^{\cW}|$, and
$\cN_L^{\cW}\subseteq S_L$ gives $\mathrm{Lip}_\bullet(f')\le L$ for every $f'\in\cN_L$.
For the fit transfer, with
$u_i\coloneqq \tau f_w(x_i)-y_i$ ($\norm{u_i}\le2$) and
$\Delta_i\coloneqq \tau f_{w'}(x_i)-\tau f_w(x_i)$ ($\norm{\Delta_i}\le\eps/16$),
\[
 \norm{u_i+\Delta_i}^2\le\norm{u_i}^2+2\norm{u_i}\norm{\Delta_i}+\norm{\Delta_i}^2
 \le\norm{u_i}^2+\frac{\eps}{4}+\frac{\eps^2}{256}
 \le \norm{u_i}^2+\frac{\eps}{2}.\qedhere
\]
\end{proof}

The same entropy calculation closes every net union bound once its exponent is at~least~$Q$.

\begin{corollary}[Net accounting]\label{cor:netsum}
Let $\eps\in(0,1]$, $\delta\in(0,1)$, let $Q = p\log\!\Big(\tfrac{48\,WJ}{\eps}\Big) + \log\Big(\tfrac{16}{\delta}\Big)$
be as in \eqref{eq:Q}, and let
$\cN$ be the truncated function net $\cN_L$ or the parameter net $\cN_L^{\cW}$ of
Lemma~\ref{lem:net}, for any $L\in(0,\infty]$. Then for every
$E\ge Q$,
\[
 |\cN|\,e^{-E}\;\le\;e^{-(E-Q)}\,\frac{\delta}{16}\;\le\;\frac{\delta}{16} .
\]
\end{corollary}
\begin{proof}
By Lemma~\ref{lem:net} and \eqref{eq:Q}, in either case
$\log|\cN|\le p\log(48WJ/\eps)=Q-\log(16/\delta)$, so
$|\cN|\,e^{-E}\le\exp\big(Q-\log\tfrac{16}{\delta}-E\big)
=e^{-(E-Q)}\,\tfrac{\delta}{16}$.
\end{proof}

The net also yields, in the converse direction, a self-contained quantitative version of
an observation of \citet{WHZ23}: below-noise fitting itself requires complexity of order
$n\eps^2$.

\begin{proposition}[Necessity of overparametrization; cf.\ {\citealp[Theorem~D.4 of arXiv:2202.11592v1]{WHZ23}}]
\label{prop:overparam}
Let Assumption~\ref{ass:model} hold and let $\cF$ admit a $(p,J,W)$-parametrization of
scalar Borel measurable functions. If $n\eps^2\ge1024\,Q$, with $Q$ as in \eqref{eq:Q}, then with
probability at least $1-3e^{-n\eps^2/2048}$, simultaneously for all $w\in\cW$:
\[
 \frac1n\sum_{i=1}^n\big(f_w(x_i)-y_i\big)^2\;>\;\sigma^2-\eps .
\]
\end{proposition}
\begin{proof}
  See Appendix~\ref{app:nonvac}.
\end{proof}

\begin{remark}[The meaningful regime]\label{rem:nonvacuous}
Proposition~\ref{prop:overparam} shows that, for classes of Borel measurable functions,
except with probability $3e^{-n\eps^2/2048}$, a below-noise fit requires
$Q>n\eps^2/1024$. We call this the
\emph{meaningful regime}. It is used only to interpret the results, not as a hypothesis.
\end{remark}

\subsection{A Master Theorem}

The two laws of this paper share a deterministic reduction based on the
noise-correlation argument in the proof of \citet[Theorem~2]{BS21}.
Theorem~\ref{thm:master} expresses this reduction in terms of six probabilistic conditions.
Conditions \emph{(S0)}--\emph{(S3)} control failures that are independent of the candidate
function, so their probabilities enter the bound only once; the bounds in
\emph{(S4)}--\emph{(S5)} may depend on the covering number. We state the theorem in the generality of the
companion paper \citep{PeterModular}, which uses its further features: vector-valued
predictors ($K>1$), a \emph{latent centering index} $(k_i)$ for mixtures, and nontrivial
events $\cA$ for mixtures and dependent data; the statement also admits an enlargement of
the probability space at slot \emph{(S0)}. In this paper $K=1$, the index is trivial and
$\cC=\cA=\Omega$.

Theorems~\ref{thm:baseline} and~\ref{thm:logconcave} take the Lipschitz threshold
$L^\star=\frac{\eps}{2^9}\sqrt{nd/(c_\bullet Q)}$ for the concentration constant
$c_\bullet$ of the theorem at hand, so the induced variance proxy is
in both cases
\begin{equation}\label{eq:vstar}
 v^\star\;\coloneqq\;\frac{c_\bullet\,(L^\star)^2}{d}\;=\;\frac{n\eps^2}{2^{18}\,Q},
\end{equation}
independently of $c_\bullet$.

\begin{theorem}[Master law of robustness]\label{thm:master}
Let $K\ge1$, let $\cF=\{f_w:w\in\cW\}$ admit a $(p,J,W)$-parametrization
(Definition~\ref{def:param}), let $\mathrm{Lip}_\bullet$ and $\tau$ be as in
Lemma~\ref{lem:net}, and suppose $\tau f_w$ is $\mathbb{B}_2^K$-valued for every $w\in\cW$. Let
$(x_i,y_i)_{i\le n}$, together with a latent index $(k_i)_{i\le n}\in \Lambda^n$, 
be defined on a common probability space, with $y_i\in \mathbb{B}_2^K$ almost surely; let $g:\R^d\to \mathbb{B}_2^K$ be
measurable, $z_i\coloneqq y_i-g(x_i)$, $\sigma^2>0$, $\eps\in(0,1]$ with
$\eps\le\sigma^2$, and $\delta\in(0,1)$. For $L^\star>0$, let $\cN\coloneqq\cN_{L^\star}$ be the truncated function net of
Lemma~\ref{lem:net}, and let $Q$ be as in \eqref{eq:Q}, so that
$\log|\cN|\le Q-\log(16/\delta)$. Put
\[
 G_1\coloneqq\Big\{\tfrac1n\sum_i\norm{z_i}^2\ge\sigma^2-\tfrac\eps8\Big\},
 \qquad
 G_3\coloneqq\Big\{\sum_i\inner{z_i}{-g(x_i)}\le\tfrac{n\eps}{16}\Big\},
\]
\[
 G_2\coloneqq\Big\{\Big|\sum_i\inner{z_i}{c_{k_i}}\Big|\le\tfrac{n\eps}{16}
 \ \text{ for every family }(c_\ell)_\ell\subset \mathbb{B}_2^K\Big\}.
\]
Suppose there are measurable events $\cA$ and $\cC$, and centering vectors
$m^{(\ell)}(f')\in \mathbb{B}_2^K$ for each $f'\in\cN$, such that, writing
$a_i(f')\coloneqq f'(x_i)-m^{(k_i)}(f')$, the following six conditions hold:
\begin{enumerate}[label=\emph{(S\arabic*)},start=0,leftmargin=3.2em,itemsep=1pt]
\item $\Prob[\cC^c]\le\delta/8$;
\item $\Prob[G_1^c\cap\cC]\le\delta/4$;
\item $\Prob[G_2^c\cap\cC]\le\delta/4$;
\item $\Prob[G_3^c\cap\cC]\le\delta/8$;
\item $\Prob[\cA^c\cap\cC]\le\delta/8$;
\item $\displaystyle\sum_{f'\in\cN}
 \Prob\Big[\Big\{\sum_i\inner{z_i}{a_i(f')}\ge\tfrac{n\eps}{16}\Big\}\cap\cC\cap\cA\Big]
 \le\delta/8$.
\end{enumerate}
Then with probability at least $1-\delta$, it holds simultaneously for all $w\in\cW$: 
\begin{equation*}
  \frac1n\sum_i\norm{f_w(x_i)-y_i}^2\le\sigma^2-\eps \quad \Longrightarrow \quad \mathrm{Lip}_\bullet(f_w)>L^\star.
\end{equation*}
\end{theorem}
\begin{proof}
\emph{Reduction to the net.} Suppose some $w\in\cW$ has
$\frac1n\sum_i\norm{f_w(x_i)-y_i}^2\le\sigma^2-\eps$ and
$\mathrm{Lip}_\bullet(f_w)\le L^\star$. Since $\norm{y_i}\le1$, the reduction $\tau$ does
not increase the loss and does not increase $\mathrm{Lip}_\bullet$
(Lemma~\ref{lem:trunc}), so
$w\in S_{L^\star}$ and $\tau f_w$ fits at level
$\sigma^2-\eps$; by Lemma~\ref{lem:net} there is $w'\in\cN_{L^\star}^{\cW}$ such that
$f'\coloneqq\tau f_{w'}\in\cN$ is $\mathbb{B}_2^K$-valued, fits at level $\sigma^2-\eps/2$, and satisfies
$\mathrm{Lip}_\bullet(f')\le L^\star$.

\emph{The pairing event is forced.} Fix $f'\in\cN$ and work on
$G_1\cap G_2\cap G_3$. The hypothesis of Lemma~\ref{lem:decomp} with
$\xi=\eps/2$ is exactly $G_1$, so a fit of $f'$ at level $\sigma^2-\eps/2$ gives
$\sum_i\inner{z_i}{f'(x_i)-g(x_i)}\ge\frac{3n\eps}{16}$. Splitting along the centering,
\[
 \sum_i\inner{z_i}{f'(x_i)-g(x_i)}
 =\sum_i\inner{z_i}{a_i(f')}
 +\sum_i\inner{z_i}{m^{(k_i)}(f')}
 +\sum_i\inner{z_i}{-g(x_i)},
\]
where the second sum is at most $\frac{n\eps}{16}$ in absolute value on $G_2$ (the family
$(m^{(\ell)}(f'))_\ell$ lies in $\mathbb{B}_2^K$) and the third is at most $\frac{n\eps}{16}$ on
$G_3$. Hence $\sum_i\inner{z_i}{a_i(f')}\ge\frac{3n\eps}{16}-\frac{2n\eps}{16}
=\frac{n\eps}{16}$.

\emph{Accounting.} Consequently the event that the conclusion fails for some $w\in\cW$ is
contained in
\[
 \cC^c\;\cup\;(G_1^c\cap\cC)\;\cup\;(G_2^c\cap\cC)\;\cup\;(G_3^c\cap\cC)
 \;\cup\;(\cA^c\cap\cC)\;\cup\;
 \bigcup_{f'\in\cN}\Big(\Big\{\sum_i\inner{z_i}{a_i(f')}\ge\tfrac{n\eps}{16}\Big\}
 \cap\cC\cap\cA\Big),
\]
whose probability is at most
$\frac\delta8+\frac\delta4+\frac\delta4+\frac\delta8+\frac\delta8+\frac\delta8=\delta$ by
(S0)--(S5).
\end{proof}

\subsection{The Baseline Theorem}

We first apply the master theorem to scalar predictors and i.i.d.\ data with a
$c$-isoperimetric covariate distribution. This recovers the
Bubeck--Sellke law, which serves as a baseline for the exponential law in
Section~\ref{sec:logconcave}. 

\begin{theorem}[Bubeck--Sellke law of robustness, quantitative variant;
{\citealp[Theorem~3]{BS21}}]\label{thm:baseline}
Let Assumption~\ref{ass:model} hold with $\mu$ $c$-isoperimetric for some $c>0$, and let
$\cF$ admit a $(p,J,W)$-parametrization of scalar functions. If
\[
 n\,\eps^{2}\;\ge\;2^{12}\,\log\!\Big(\frac{16}{\delta}\Big),
\]
then with probability at least $1-\delta$, simultaneously for all $w\in\cW$:
\[
 \frac1n\sum_{i=1}^n\big(f_w(x_i)-y_i\big)^2\;\le\;\sigma^2-\eps
 \qquad\Longrightarrow\qquad
 \Lip(f_w)\;\ge\;\frac{\eps}{2^{9}}\,\sqrt{\frac{n\,d}{c\,Q}}\;\eqqcolon\;L^\star .
\]
\end{theorem}

\begin{proof}
We verify the slots of Theorem~\ref{thm:master} with $K=1$,
$\mathrm{Lip}_\bullet=\Lip$, $\tau$ the clipping to $[-1,1]$, trivial latent index
$k_i\equiv1$, and $\cC=\cA=\Omega$ the full space, so that \emph{(S0)} and \emph{(S4)} hold trivially. 
Let $\cN\coloneqq\cN_{L^\star}$ and
$v^\star\coloneqq c(L^\star)^2/d=\frac{n\eps^2}{2^{18}Q}$, and let $m^{(1)}(f')\coloneqq\E_\mu f'\in[-1,1]$, 
so that $a_i(f')=f'(x_i)-\E_\mu f'$. Since there is only one latent index, we get
$G_2=\{|\sum_iz_i|\le\frac{n\eps}{16}\}$. Throughout we use the
sample condition in the form $n\eps^2\ge2^{12}\log(16/\delta)$.

\emph{(S1).} By Lemma~\ref{lem:noise} with $\xi=\eps/2$, we get $\Prob[G_1^c]\le e^{-n\eps^2/576}\le(\delta/16)^{7}\le\delta/4$.

\emph{(S2) and (S3).} Conditionally on the covariates the $z_i$ are independent, centered
with values in $[-2,2]$, so Lemma~\ref{lem:globalsums} (with $\cG_0=\sigma(x_{1:n})$) gives
$\Prob[G_2^c]\le2e^{-n\eps^2/2048}\le2(\delta/16)^{2}\le\delta/4$ and
$\Prob[G_3^c]\le e^{-n\eps^2/2048}\le(\delta/16)^{2}\le\delta/8$.

\emph{(S5).} Each $f'\in\cN$ is $[-1,1]$-valued with $\Lip(f')\le L^\star$. If
$\Lip(f')=0$, then $f'$ is constant, so $a_i(f')=0$ for every $i$ and the event
$\{\sum_iz_ia_i(f')\ge\frac{n\eps}{16}\}$ is empty; its probability $0$ is at most
$e^{-11Q}$. Otherwise $v\coloneqq c\Lip(f')^2/d\in(0,v^\star]$, and by
$c$-isoperimetry $a_i(f')\in\expTail^{2}(1,\sqrt{2v})$. Corollary~\ref{cor:pairing} with
$s=\frac{n\eps}{16}$ gives
\[
 \Prob\Big[\sum_iz_ia_i(f')\ge\frac{n\eps}{16}\Big]
 \;\le\;\exp\Big(-\frac{(n\eps/16)^2}{88\,n\,v}\Big)
 \;\le\;\exp\Big(-\frac{n\eps^2}{22528\,v^\star}\Big)
 \;=\;e^{-\frac{2^{18}}{22528}Q}\;\le\;e^{-11Q}.
\]
In either case the probability is at most $e^{-11Q}$, so by Corollary~\ref{cor:netsum}
with $E=11Q$,
\[
 \sum_{f'\in\cN}\Prob\Big[\sum_iz_ia_i(f')\ge\frac{n\eps}{16}\Big]
 \;\le\;|\cN|\,e^{-11Q}\;\le\;\frac{\delta}{16}\;\le\;\frac\delta8 .
\]

Theorem~\ref{thm:master} therefore applies and gives, with probability at least $1-\delta$,
that every $w\in\cW$ fitting at level $\sigma^2-\eps$ has $\Lip(f_w)>L^\star$, hence
$\Lip(f_w)\ge L^\star$.
\end{proof}

\begin{remark}[Measurability of the uniform event]\label{rem:measurable}
Put $a\coloneqq\sigma^2-\eps$ and $R_f\coloneqq n^{-1}\sum_i(f(x_i)-y_i)^2$. The
conclusion of Theorem~\ref{thm:baseline} concerns the intersection
\[
 \mathcal U\coloneqq\bigcap_{w\in\cW}
 \{R_{f_w}\le a\Longrightarrow\Lip(f_w)\ge L^\star\}
\]
over the uncountable index set $\cW$. This event might not be measurable. Each proof,
here and in every later theorem, constructs a \emph{measurable}
event $H\subseteq\mathcal U$ with $\Prob[H]\ge1-\delta$. In the notation of
Theorem~\ref{thm:master}, the accounting step of its proof shows this for the finite
intersection of measurable events
\[
 H\coloneqq\cC\cap\cA\cap G_1\cap G_2\cap G_3\cap
 \bigcap_{f'\in\cN}\Big\{\sum_i\inner{z_i}{a_i(f')}<\tfrac{n\eps}{16}\Big\}.
\]
Measurability of $\mathcal U$ itself is neither claimed nor needed.
Appendix~\ref{app:nonvac} shows that $\mathcal U$ is Borel whenever $\cW$ is
$\sigma$-compact.
\end{remark}

\begin{remark}[Baseline specialization and endpoint conventions]\label{rem:baselineform}
Theorem~\ref{thm:baseline} has the rate of \citet[Theorem~3]{BS21} with different
explicit constants. Their bound is
$\Lip\ge\frac{\eps}{2^9\sqrt c}\sqrt{nd/(p\log(60WJ/\eps)+\log(4/\delta))}$ under
$9^4\log(8/\delta)\le n\eps^2$; ours is $\Lip\ge\frac{\eps}{2^9}\sqrt{nd/(cQ)}$ under
$n\eps^2\ge2^{12}\log(16/\delta)$. Neither set of absolute constants is claimed optimal.
\end{remark}

\section{A Law of Robustness under Exponential Concentration}\label{sec:logconcave}

A Poincar\'e inequality yields only \emph{exponential} concentration of Lipschitz
functions, so general isotropic log-concave covariates do not satisfy \eqref{eq:isop} with
bounded $c$ (the product Laplace law of Proposition~\ref{rem:lcnoconverse} is the standard
obstruction) and lie outside
the reach of Theorem~\ref{thm:baseline}. Theorem~\ref{thm:logconcave} retains the
unchanged rate and absolute constant under this weaker tail, at the price of the burn-in
$n\ge Q/61$. Without the burn-in, the law still holds, for $n\le Q$ at the level
$4\sqrt{n/Q}\,L^\star$ or better (Proposition~\ref{prop:lcbeyond}). For $Q>2^{28}n$,
Proposition~\ref{prop:lcburnin} exhibits an instance where this level is attained up to
an absolute factor. Corollary~\ref{cor:logconcave} combines the
theorem with Klartag's bound on the KLS constant: for full-dimensional log-concave laws
satisfying its covariance normalization, the resulting constant loses a factor $\log d$,
which the recently announced proof of the KLS conjecture removes (part~(ii)).
The source of the burn-in is that a $\psi_1$ fluctuation admits a quadratic upper bound 
on its log-MGF only over a bounded range, so the Chernoff
optimization behind the pairing bound may hit the boundary of that range after the net
union bound; the burn-in makes this boundary branch strong enough to
absorb the net factor.

\begin{definition}[$c$-exponential concentration]\label{def:expconc}
A probability measure $\mu$ on $\R^d$ satisfies \emph{$c$-exponential
concentration} if for every bounded $L$-Lipschitz $f:\R^d\to\R$ and every
$t\ge0$,
\begin{equation}\label{eq:expconc}
 \Prob_{X\sim\mu}\big[|f(X)-\E_\mu f|\ge t\big]\;\le\;
 2\exp\!\Big(-\frac{\sqrt d\,t}{\sqrt c\,L}\Big).
\end{equation}
The boundary case $c=0$ is read as the Dirac convention of
Definition~\ref{def:isop}. We attach to $f$ the scale proxy $\sqrt v$
with $v\coloneqq cL^2/d$, the \emph{same} quantity $v$ as in the Gaussian
case, so that all bookkeeping below is directly comparable to
Section~\ref{sec:skeleton}.
\end{definition}

\begin{remark}[Relation to Definition~\ref{def:isop}]\label{rem:lcvsgauss}
Note that $c$-isoperimetry implies $4c$-exponential concentration: for $t\ge\sqrt v$
one has $t^2/(2v)\ge t/(2\sqrt v)$, while for $t<\sqrt v$ the right-hand
side of \eqref{eq:expconc} at scale $2\sqrt v$ exceeds $2e^{-1/2}>1$.
Thus Theorem~\ref{thm:logconcave} below covers the setting of
Theorem~\ref{thm:baseline} as well (with $c$ inflated by $4$ and the
extra burn-in $n\ge Q/61$ as the price); Theorem~\ref{thm:baseline} remains
the stronger statement where it applies, since it carries no burn-in.
\end{remark}

The converse fails even under the covariance normalization of
Corollary~\ref{cor:logconcave}; Appendix~\ref{app:lc} gives a product-Laplace
counterexample (Proposition~\ref{rem:lcnoconverse}).

The following self-contained fact converts a Poincar\'e inequality into
\eqref{eq:expconc} with explicit constants. The statement is classical and goes back to 
Gromov--Milman \citeyearpar[Theorem~4.1]{GM83}. The
exponential-moment bound obtained by applying the Poincar\'e inequality
to $e^{\lambda f/2}$ and iterating -- the argument we follow below --
is due to Bobkov--Ledoux \citeyearpar[Proposition~4.1]{BL97}. 
A metric-measure form is established by Ledoux \citeyearpar[Corollary~3.2, Eq.~(1.9)]{Ledoux01}.

\begin{proposition}[Poincar\'e implies exponential concentration]\label{prop:poincexp}
For locally Lipschitz $g:\R^d\to\R$ let
\[
 |\nabla g|(x)\;\coloneqq\;\limsup_{y\to x}\ \frac{|g(y)-g(x)|}{\norm{y-x}}
\]
denote its \emph{local slope}. Suppose $\mu$ satisfies the ($d$-normalized) Poincar\'e
inequality
\[
 \Var_\mu[g]\;\le\;\frac{c_0}{d}\,\E_\mu\big[\,|\nabla g|^2\,\big]
 \qquad\text{for every bounded Lipschitz }g.
\]
Then $\mu$ satisfies $9c_0$-exponential concentration.
\end{proposition}
\begin{proof}
  See Appendix~\ref{app:lc}.
\end{proof}

Here $\Var_\mu[g]\coloneqq\inf_{a\in\R}\E_\mu[(g-a)^2]\in[0,\infty]$, so the
hypothesis is meaningful for every locally Lipschitz $g$. 
The local slope is Borel measurable. When $\mu\ll dx$, Rademacher's theorem 
\citep[Section~3.1.2]{EvansGariepy15} gives
$|\nabla g|=\norm{\nabla g}$ $\mu$-a.e., so the hypothesis agrees with the usual
Euclidean-gradient Poincar\'e inequality; this covers every application below. For
singular $\mu$, the local-slope form is the intrinsic formulation. Appendix~\ref{app:lc}
gives the proof, which uses the inequality only for bounded Lipschitz functions and
requires no differentiability.

The next lemma replaces the all-$\lambda$ product MGF of Lemma~\ref{lem:key} by its
$\alpha=1$ capped counterpart. The cap is intrinsic (Remark~\ref{rem:lccap}) and is
what produces the burn-in in Theorem~\ref{thm:logconcave}.

\begin{lemma}[Capped product MGF]\label{lem:lckey}
Let $A\in\expTail^{1}(1,\sqrt v)$ (Definition~\ref{def:exptail}) for
some $v>0$, and let $Z$ satisfy $|Z|\le2$ a.s.\ and
$\E[Z\,|\,A]=0$. Then
\[
 \E\,e^{\lambda ZA}\;\le\;\exp\!\big(16\,\lambda^2v\big)
 \qquad\text{for all }|\lambda|\le\frac1{4\sqrt v}.
\]
(For $v=0$, i.e.\ $A=0$ a.s., the bound holds trivially.)
\end{lemma}
\begin{proof}

This is Lemma~\ref{lem:keycap} with $K=1$, $\alpha=1$ and $s=\sqrt v$, whereby Lemma~\ref{lem:psimom}
ensures that $A$ satisfies the necessary moment condition.
In this case $\eta_\alpha=0$, so the constant is
$2^{4-\eta_\alpha}=16$ and the range is $|\lambda|\le3^{\eta_\alpha}/(4s)=1/(4\sqrt v)$.
\end{proof}

\begin{remark}[The cap is intrinsic]\label{rem:lccap}
The cap in Lemma~\ref{lem:lckey} is unavoidable. If $Z$ is an independent Rademacher
sign and $A$ is centered Laplace with scale $\sqrt v$, then $ZA\overset{d}{=}A$ and
\[
 \E\,e^{\lambda ZA}
 =
 \frac{1}{1-\lambda^2v},
 \qquad
 |\lambda|<\frac1{\sqrt v},
\]
while the moment generating function is infinite outside this interval. Hence only a 
bounded-range quadratic bound on the log-MGF, and therefore only a two-branch Chernoff bound
(Lemma~\ref{lem:cappedtail}), is possible; the burn-in lets the linear branch absorb
the net.
\end{remark}

The two-branch tail bound, instantiated at the exponential scale, reads as follows; the
boundary between its branches is what will generate the burn-in of
Theorem~\ref{thm:logconcave}.

\begin{corollary}[Exponential pairing tail]\label{cor:lcpairing}
In the setting of Assumption~\ref{ass:model}, let $\alpha:\R^d\to\R$ be
measurable with $\alpha(X)\in\expTail^{1}(1,\sqrt v)$ for $X\sim\mu$,
and set $A_i=\alpha(X_i)$. Then for every $s\ge0$,
\[
 \Prob\Big[\sum_{i=1}^nz_iA_i\ge s\Big]\;\le\;
 \begin{cases}
 \exp\!\Big(-\dfrac{s^2}{64\,n\,v}\Big), & \text{if }s\le8n\sqrt v,\\[8pt]
 \exp\!\Big(n-\dfrac{s}{4\sqrt v}\Big), & \text{always.}
 \end{cases}
\]
\end{corollary}
\begin{proof}
The pairs $(z_i,A_i)$ are i.i.d., $|z_i|\le2$ and
$\E[z_i\,|\,A_i]=\E[\E[z_i|x_i]\,|\,A_i]=0$, so by
Lemma~\ref{lem:lckey}, $\E e^{\lambda\sum_iz_iA_i}\le e^{16n\lambda^2v}$
for $0\le\lambda\le\frac1{4\sqrt v}$. Both branches now follow from 
Lemma~\ref{lem:cappedtail} with $a=16nv$ and $\theta_0=\frac1{4\sqrt v}$. Then
$2a\theta_0=8n\sqrt v$, $\frac{s^2}{4a}=\frac{s^2}{64nv}$, and
$a\theta_0^2-\theta_0 s=\frac{16nv}{16v}-\frac{s}{4\sqrt v}=n-\frac{s}{4\sqrt v}$.
\end{proof}

We now verify the master theorem with the exponential pairing tail in place of the
Gaussian one. Every function-independent slot is unchanged; only the net-indexed pairing
slot sees the cap on the moment-generating function, and the condition $n\ge Q/61$
lets its boundary branch beat the complexity $Q$.

\begin{theorem}[Law of robustness under exponential concentration]
\label{thm:logconcave}
Let Assumption~\ref{ass:model} hold with $\mu$ satisfying
$c$-exponential concentration for some $c>0$, and let $\cF$ admit a
$(p,J,W)$-parametrization of scalar functions. If
\[
 n\,\eps^{2}\;\ge\;2^{12}\,\log\!\Big(\frac{16}{\delta}\Big)
 \qquad\text{and}\qquad
 n\;\ge\;\frac Q{61},
\]
with $Q$ as in \eqref{eq:Q}, then with probability at least $1-\delta$,
simultaneously for all $w\in\cW$:
\[
 \frac1n\sum_{i=1}^n\big(f_w(x_i)-y_i\big)^2\;\le\;\sigma^2-\eps
 \qquad\Longrightarrow\qquad
 \Lip(f_w)\;\ge\;\frac{\eps}{2^{9}}\,\sqrt{\frac{n\,d}{c\,Q}}\;\eqqcolon\;L^\star .
\]
\end{theorem}

\begin{proof}
We verify the slots of Theorem~\ref{thm:master} with $K=1$,
$\mathrm{Lip}_\bullet=\Lip$, $\tau$ the clipping to $[-1,1]$, trivial latent index
$k_i\equiv1$, and $\cC=\cA=\Omega$ the full space, so that \emph{(S0)} and \emph{(S4)} hold with
zero failure probability. Let $\cN\coloneqq\cN_{L^\star}$ denote the net of Lemma~\ref{lem:net}, and
$v^\star\coloneqq c(L^\star)^2/d=\frac{n\eps^2}{2^{18}Q}$, and take the single centering
vector $m^{(1)}(f')\coloneqq\E_\mu f'\in[-1,1]$, so that $a_i(f')=f'(x_i)-\E_\mu f'$;
since the index is trivial, we have again $G_2=\{|\sum_iz_i|\le\frac{n\eps}{16}\}$.

\emph{(S1)--(S3).} These slots use only the boundedness $|z_i|\le2$, $|g(x_i)|\le1$, the
independence of the observations and the conditional centering $\E[z_i\mid x_i]=0$---never the concentration properties of the covariate distribution---and the first sample
condition is that of Theorem~\ref{thm:baseline}, so the verification given there applies
verbatim.

\emph{(S5).} Each $f'\in\cN$ is $[-1,1]$-valued with $\Lip(f')\le L^\star$. By
\(c\)-exponential concentration, $\alpha\coloneqq f'-\E_\mu f'$
satisfies the tail hypothesis of Corollary~\ref{cor:lcpairing} with scale parameter
$
 v(f')
 \coloneqq
 \frac{c\,\Lip(f')^2}{d}
 \le
 v^\star .
$
We partition the net deterministically into
\[
 \cN_{\mathrm{sub}}
 \coloneqq
 \left\{
 f'\in\cN:
 v(f')\ge\frac{\eps^2}{2^{14}}
 \right\},
\]
and its complement
\[
 \cN_{\mathrm{bd}}
 \coloneqq
 \left\{
 f'\in\cN:
 v(f')<\frac{\eps^2}{2^{14}}
 \right\}.
\]

\emph{Sub-Gaussian branch.} Let \(f'\in\cN_{\mathrm{sub}}\), and set
\(s\coloneqq\frac{n\eps}{16}\). The validity condition \(s\le8n\sqrt{v(f')}\) for the
first branch of Corollary~\ref{cor:lcpairing} holds, since \(v(f')\ge\eps^2/2^{14}\).
Therefore
\[
\begin{aligned}
 \Prob\!\left[
 \sum_{i=1}^n z_ia_i(f')
 \ge
 \frac{n\eps}{16}
 \right]
 &\le
 \exp\!\left(
 -\frac{(n\eps/16)^2}{64\,n\,v(f')}
 \right)\\
 &=
 \exp\!\left(
 -\frac{n\eps^2}{2^{14}v(f')}
 \right)\\
 &\le
 \exp\!\left(
 -\frac{n\eps^2}{2^{14}v^\star}
 \right)\\
 &=
 e^{-16Q},
\end{aligned}
\]
where the second inequality uses \(v(f')\le v^\star\).

\emph{Boundary branch.}
Now let \(f'\in\cN_{\mathrm{bd}}\). If \(v(f')=0\), then
\(a_i(f')=0\) almost surely, so the pairing probability is \(0\).

Otherwise, by the definition of \(\cN_{\mathrm{bd}}\),
$0<v(f')<\frac{\eps^2}{2^{14}}$, so $0<\sqrt{v(f')}<\frac{\eps}{128}$.
The second branch of Corollary~\ref{cor:lcpairing} carries no restriction on $s$, and we
record the two consequences we shall need. First, using only
$\sqrt{v(f')}<\eps/128$,
\[
 \Prob\!\left[
 \sum_{i=1}^n z_ia_i(f')
 \ge
 \frac{n\eps}{16}
 \right]
 \le
 \exp\!\left(
 n-\frac{n\eps}{64\sqrt{v(f')}}
 \right)
 <
 e^{-n}\,.
\]
Second, using that $\Lip(f')\leq L^*=\frac{\eps}{2^{9}}\,\sqrt{\frac{n\,d}{c\,Q}}$, 
we obtain $v(f')\le v^\star=\frac{n\eps^2}{2^{18}Q}$ and therefore
\[
 \frac{n\eps}{64\sqrt{v(f')}}
 \;\ge\;
 \frac{n\eps\,2^{9}\sqrt Q}{64\,\eps\sqrt n}
 \;=\;
 8\sqrt{nQ},
 \quad\text{whence}\quad
 \Prob\!\left[\sum_{i=1}^n z_ia_i(f')\ge\frac{n\eps}{16}\right]
 \;\le\;
 \exp\!\left(n-8\sqrt{nQ}\right).
\]
Now we apply Corollary~\ref{cor:netsum} to ensure that the net sum of (S5) is bounded by $\delta/8$.
For the sub-Gaussian branch, with
$E=16Q$, we get
\[
 |\cN|e^{-16Q} \;\le\; \frac{\delta}{16}\,.
\]
For the boundary branch, we get the following two bounds for $E=8\sqrt{nQ}-n$ and $E=n$, respectively:
\begin{align*}
  |\cN|\exp\!\left(n-8\sqrt{nQ}\right)
  &\;\le\;
  \frac{\delta}{16}
  \qquad\text{provided } n+Q\le8\sqrt{nQ},\\
  |\cN|e^{-n}
  &\;\le\;
  \frac{\delta}{16}
  \qquad\text{provided } n\ge Q .
\end{align*}
Note that the first condition is equivalent to $ \big(4-\sqrt{15}\big)^{2}\,Q\;\le\;n\;\le\;\big(4+\sqrt{15}\big)^{2}\,Q$, where 
$\big(4-\sqrt{15}\big)^{2}=0.016133\ldots,\quad\big(4+\sqrt{15}\big)^{2}=61.98\ldots$.
The union of the two admissible ranges is
$\{n\ge(4-\sqrt{15})^2Q\}$, which contains $\{n\ge Q/61\}$ because
$1/61=0.016393\ldots>(4-\sqrt{15})^2$. The burn-in hypothesis $n\ge Q/61$ therefore
ensures the bound for the boundary branch, and this is the only place where it is used. Since both
estimates bound the same probability, every $f'\in\cN_{\mathrm{bd}}$ has pairing
probability at most $\min\big(e^{-n},\,e^{\,n-8\sqrt{nQ}}\big)$. Splitting the sum
over the net along the partition gives
\[
 \sum_{f'\in\cN}
 \Prob\!\left[\sum_{i=1}^n z_ia_i(f')\ge\frac{n\eps}{16}\right]
 \;\le\;
 |\cN|\,e^{-16Q}
 +
 |\cN|\min\!\big(e^{-n},\,e^{\,n-8\sqrt{nQ}}\big)
 \;\le\;
 \frac{\delta}{16}+\frac{\delta}{16}
 \;=\;
 \frac{\delta}{8},
\]
which verifies \emph{(S5)}.
Theorem~\ref{thm:master} now gives the conclusion, exactly as in the closing step of
Theorem~\ref{thm:baseline}.
\end{proof}

The burn-in of Theorem~\ref{thm:logconcave} is used only when the capped Chernoff bound
enters its linear branch. The next construction shows that some restriction on $Q/n$ is
genuinely necessary, although it does not claim that $Q/61$ is the sharp transition.

\begin{proposition}[Necessity of the burn-in]
\label{prop:lcburnin}
Let $n\in\mathbb N$, $\eps\in(0,1]$, and
$\delta\in(0,\tfrac12)$ satisfy $n\eps^2\ge 2^{12}\log(16/\delta)$,
and let $\vartheta\ge 2^{27}$. Then there exist a dimension $d$, a
product probability measure $\mu$ on $\R^d$ satisfying
$c$-exponential concentration with $c=36$, a label law such that
$(x_i,y_i)_{i\le n}$ satisfies Assumption~\ref{ass:model} with
$g\equiv0$ and $\sigma^2=1$, and a class
$\cF=\{f_w:w\in\cW\}$ of $M=\big\lceil e^{2\vartheta n}\big\rceil$
functions admitting a $(1,J,1)$-parametrization, whose complexity term
\eqref{eq:Q} satisfies
\[
 2\vartheta n<Q\le (2\vartheta+1)n;
\]
in particular $n<Q/(2\vartheta)\le Q/2^{28}$, so the burn-in
$n\ge Q/61$ of Theorem~\ref{thm:logconcave} is violated.
Moreover:
\begin{enumerate}[label=(\roman*)]
\item every $w\in\cW$ satisfies
\[
 \Lip(f_w)
 \le
 18\,\eps\sqrt{\frac{nd}{cQ}}\sqrt{\frac nQ}
 <
 \frac{\eps}{2^9}\sqrt{\frac{nd}{cQ}}
 =L^\star;
\]
\item with probability at least
\[
 1-\exp\!\big(-e^{n(\vartheta-\log2)}\big)
 \ge
 1-\frac{\delta}{16},
\]
some $w\in\cW$ has empirical squared error exactly
$\sigma^2-\eps$.
\end{enumerate}
Thus every hypothesis of Theorem~\ref{thm:logconcave} except the
burn-in is satisfied, while its conclusion fails with probability at
least $1-\delta/16>\delta$. In particular, the burn-in cannot be
removed from Theorem~\ref{thm:logconcave}.
\end{proposition}

\begin{proof}
  See Appendix~\ref{app:lcburnin}.
\end{proof}

The counterexample in Proposition~\ref{prop:lcburnin} has normalized log-concave covariates
as well: its product-Laplace distribution satisfies $\mathrm{Cov}(\mu)=\frac2d\Id_d$. 
Thus, for the threshold $L^\star$, some restriction on $Q/n$ remains necessary.

Without the burn-in, the argument of Theorem~\ref{thm:logconcave} still yields a law of
robustness, at a Lipschitz threshold that for $Q\gg n$ is smaller than $L^\star$ by a
factor of order $\sqrt{n/Q}$. Its proof uses only the linear branch of
Corollary~\ref{cor:lcpairing}.

\begin{proposition}[Law of robustness without burn-in]
\label{prop:lcbeyond}
Let Assumption~\ref{ass:model} hold with $\mu$ satisfying
$c$-exponential concentration for some $c>0$, and let $\cF$ admit a
$(p,J,W)$-parametrization of scalar functions. If
\[
 n\,\eps^{2}\;\ge\;2^{12}\,\log\!\Big(\frac{16}{\delta}\Big),
\]
then with probability at least $1-\delta$, simultaneously for all $w\in\cW$:
\[
 \frac1n\sum_{i=1}^n\big(f_w(x_i)-y_i\big)^2\;\le\;\sigma^2-\eps
 \qquad\Longrightarrow\qquad
 \Lip(f_w)\;\ge\;\frac{\eps}{2^{6}}\,\sqrt{\frac{d}{c}}\,\frac{n}{n+Q}
 \;\eqqcolon\;L^\circ,
\]
with $Q$ as in \eqref{eq:Q}. In particular, if $n\le Q$, then
$L^\circ\ge\frac{\eps}{2^{7}}\sqrt{\frac dc}\,\frac nQ=4\sqrt{\frac nQ}\,L^\star$, with
$L^\star$ as in Theorem~\ref{thm:logconcave}.
\end{proposition}

\begin{proof}
We verify the slots of Theorem~\ref{thm:master} with the threshold $L^\circ$ in place of
$L^\star$ and otherwise the choices of the proof of Theorem~\ref{thm:logconcave}: $K=1$,
$\mathrm{Lip}_\bullet=\Lip$, $\tau$ the clipping to $[-1,1]$, $k_i\equiv1$,
$\cC=\cA=\Omega$, and $m^{(1)}(f')\coloneqq\E_\mu f'$, now with
$\cN\coloneqq\cN_{L^\circ}$. Then \emph{(S0)} and \emph{(S4)} hold with zero failure
probability. The verification of \emph{(S1)--(S3)} in Theorem~\ref{thm:baseline} uses
neither the threshold nor the net, only the first sample condition, so it applies
verbatim.

\emph{(S5).} Let $f'\in\cN$ and $v(f')\coloneqq c\Lip(f')^2/d$. If $v(f')=0$, then
$a_i(f')=0$ almost surely, so the pairing probability is $0$. Otherwise, as in the proof
of Theorem~\ref{thm:logconcave}, $\alpha\coloneqq f'-\E_\mu f'$ satisfies the tail
hypothesis of Corollary~\ref{cor:lcpairing} with scale parameter $v(f')$, and
$\Lip(f')\le L^\circ$ gives
\[
 \sqrt{v(f')}\;\le\;\sqrt{\frac cd}\,L^\circ\;=\;\frac{\eps\,n}{2^{6}\,(n+Q)}.
\]
We use only the second branch of Corollary~\ref{cor:lcpairing}, which carries no
restriction on $s$. With $s=\frac{n\eps}{16}$ it gives
\[
 \Prob\!\left[\sum_{i=1}^n z_ia_i(f')\ge\frac{n\eps}{16}\right]
 \;\le\;
 \exp\!\left(n-\frac{n\eps}{64\sqrt{v(f')}}\right)
 \;\le\;
 \exp\!\left(n-\frac{n\eps\,2^{6}(n+Q)}{64\,\eps\,n}\right)
 \;=\;
 e^{-Q}.
\]
By Corollary~\ref{cor:netsum} with $E=Q$,
\[
 \sum_{f'\in\cN}
 \Prob\!\left[\sum_{i=1}^n z_ia_i(f')\ge\frac{n\eps}{16}\right]
 \;\le\;
 |\cN|\,e^{-Q}
 \;\le\;
 \frac{\delta}{16}
 \;\le\;
 \frac{\delta}{8},
\]
which verifies \emph{(S5)}. Theorem~\ref{thm:master} now gives, with probability at least
$1-\delta$, that every $w\in\cW$ fitting at level $\sigma^2-\eps$ has
$\Lip(f_w)>L^\circ$, hence $\Lip(f_w)\ge L^\circ$. Finally, if $n\le Q$, then
$n+Q\le2Q$, so $L^\circ\ge\frac{\eps}{2^{7}}\sqrt{\frac dc}\,\frac nQ$, and
$4\sqrt{n/Q}\,L^\star=\frac{\eps}{2^{7}}\sqrt{\frac{nd}{cQ}}\sqrt{\frac nQ}
=\frac{\eps}{2^{7}}\sqrt{\frac dc}\,\frac nQ$.
\end{proof}

\begin{remark}[Sharpness of the factor $n/Q$]\label{rem:lcsharp}
The instance of Proposition~\ref{prop:lcburnin} satisfies the hypotheses of
Proposition~\ref{prop:lcbeyond}, and the two conclusions hold simultaneously
with probability at least $1-\delta-\delta/16>0$. Since $Q>n$, on this
event every below-noise fit satisfies
\[
 \Lip(f_w)\ge\frac{\eps}{2^7}\sqrt{\frac dc}\,\frac nQ,
\]
while some such fit satisfies
\[
 \Lip(f_w)\le18\,\eps\sqrt{\frac dc}\,\frac nQ.
\]
Thus the dependence on $n/Q$ in Proposition~\ref{prop:lcbeyond} is optimal
up to an absolute factor for the large ratios $Q/n$ realized by the
construction. This is a worst-case statement over $c$-exponentially
concentrated covariates: for $c$-isoperimetric covariates,
Theorem~\ref{thm:baseline} retains the threshold $L^\star$ at every ratio
$Q/n$. These results do not determine whether the threshold $L^\star$
remains valid in the intermediate range $61<Q/n\le2^{28}$.
\end{remark}

To specialize Theorem~\ref{thm:logconcave} to log-concave covariates, we combine
Proposition~\ref{prop:poincexp} with an affine reduction to isotropic position and
Klartag's bound on the KLS constant. The covariance normalization controls the affine
distortion; Klartag's estimate is the sole source of the $\log d$ factor, and part~(ii)
replaces it by the announced universal bound of \citet{BKL26}.

\begin{corollary}[Law of robustness for log-concave covariates]
\label{cor:logconcave}
Let $d\ge2$, let $\mu$ be a full-dimensional log-concave probability
measure on $\R^d$ with $\norm{\mathrm{Cov}(\mu)}_{\mathrm{op}}\le\kappa/d$
(the normalization under which $N(0,\frac1d\Id_d)$ has $\kappa=1$), and
let Assumption~\ref{ass:model} and the $(p,J,W)$-parametrization hold.
\begin{enumerate}[label=(\roman*)]
\item There is an absolute constant $C_0$ (all constants are explicit except for the
universal constant in Klartag's estimate \citealp{Klartag23}) such that
$\mu$ satisfies $c$-exponential concentration with
$c\le C_0\,\kappa\log d$, and consequently, under
$n\eps^2\ge2^{12}\log(16/\delta)$ and $n\ge Q/61$, with probability
$\ge1-\delta$ every below-noise fit satisfies
\[
 \Lip(f_w)\;\ge\;\frac{\eps}{2^{9}}\,
 \sqrt{\frac{n\,d}{C_0\,\kappa\,\log d\;Q}}\,.
\]
\item Granting the universal Poincar\'e bound of \citet[Theorem~1.1]{BKL26}, with its
universal constant $C_{\mathrm{BKL}}$ (Remark~\ref{rem:kls}), $\mu$ satisfies
$c$-exponential concentration with $c\le9\,C_{\mathrm{BKL}}\,\kappa$, and consequently,
under the same two conditions, with probability $\ge1-\delta$ every below-noise fit
satisfies
\[
 \Lip(f_w)\;\ge\;\frac{\eps}{2^{9}}\,
 \sqrt{\frac{n\,d}{9\,C_{\mathrm{BKL}}\,\kappa\;Q}}\,.
\]
\end{enumerate}
\end{corollary}

\begin{proof}
  See Appendix~\ref{app:lc}.
\end{proof}

In outline, $\mu$ is the affine image $m+\Sigma^{1/2}\xi$ of an isotropic log-concave
measure $\xi$, and the change of variables gives
$C_P(\mu)\le\norm\Sigma_{\mathrm{op}}\,C_P(\xi)$. Klartag's estimate bounds $C_P(\xi)$
by a constant multiple of $\log d$, so $\norm\Sigma_{\mathrm{op}}\le\kappa/d$ yields
the $d$-normalized Poincar\'e inequality of Proposition~\ref{prop:poincexp} with
$c_0=C_{\mathrm{KLS}}\kappa\log d$ for an absolute constant $C_{\mathrm{KLS}}$. That proposition converts it into
$c$-exponential concentration with $c=9c_0$, and Theorem~\ref{thm:logconcave} gives the
stated bound. The full proof, including the covariance algebra and the conversion of
Klartag's isoperimetric estimate into a Poincar\'e bound, is given in
Appendix~\ref{app:lc}.

\begin{remark}[Recently announced proofs of the KLS conjecture]
\label{rem:kls}
Two preprints by \citet{BKL26} and \citet{SongZhang26} claim proofs of the KLS
conjecture. The arguments share a key ingredient: a criterion introduced by Song and
Zhang, whose proof is also given in \citet[Section~3]{BKL26}.

For Corollary~\ref{cor:logconcave}(ii), we use the formulation of
\citet[Theorem~1.1]{BKL26}: there is a universal constant $C_{\mathrm{BKL}}$ such that
$C_P(\mu)\le C_{\mathrm{BKL}}$ for every isotropic log-concave probability measure
in every dimension. Equivalently, an affine change of variables gives
$C_P(\mu)\le C_{\mathrm{BKL}}\norm{\mathrm{Cov}(\mu)}_{\mathrm{op}}$ for every
log-concave $\mu$ \citep[Eq.~(6)]{BKL26}. Their definition of the Poincar\'e constant
uses locally Lipschitz functions and the Euclidean gradient, as ours does.

Both proof claims appeared in October 2026 and remain at the preprint stage;
$C_{\mathrm{BKL}}$ is universal but not explicit. We therefore retain
Corollary~\ref{cor:logconcave}(i), which relies on published results, and state
part~(ii) conditional on the theorem of \citet{BKL26}.
\end{remark}

\begin{remark}[Explicit dimension-free constants for subclasses]
\label{rem:nologd}
For several subclasses, established Poincar\'e inequalities remove the $\log d$ factor
in Corollary~\ref{cor:logconcave}(i) and give explicit dimension-free constants,
independently of either KLS preprint discussed in Remark~\ref{rem:kls}.
For these subclasses, the same proof gives
$c\le9\,C_P(\xi)\,\kappa$ with explicit $C_P(\xi)$: products of one-dimensional
log-concave laws ($C_P(\xi)\le12$), strongly log-concave laws ($C_P(\xi)\le1/\rho$), and
uniform laws on Euclidean balls and ellipsoids ($C_P(\xi)\le13$).
Appendix~\ref{app:lc} records three subclasses with precise citations.
\end{remark}

The same capped argument interpolates between the $\psi_1$ and $\psi_2$ endpoints; no
new slot verification is required.

\begin{corollary}[The full $\psi_\alpha$ scale]
\label{rem:lcpsialpha}
Definition~\ref{def:expconc} is the case $\alpha=1$ of the hypothesis
\begin{equation}\label{eq:psialphaconc}
 \Prob\big[|f(X)-\E_\mu f|\ge t\big]\;\le\;
 2\exp\!\left[-\left(\frac{\sqrt d\,t}{\sqrt c\,L}\right)^{\!\alpha}\right]
 \qquad(t\ge0),
\end{equation}
that is, $f-\E_\mu f\in\expTail^{\alpha}(1,\sqrt v)$ with $v=cL^2/d$, for every bounded
$L$-Lipschitz $f$. For $\alpha\in[1,2]$ this is the Lata{\l}a--Oleszkiewicz scale
\citep{LO00} interpolating between the Poincar\'e and the logarithmic Sobolev regimes.
\emph{Theorem~\ref{thm:logconcave} holds verbatim under \eqref{eq:psialphaconc} for every
$\alpha\in[1,2]$}: same rate, same absolute constant, and the same two sample conditions.

\end{corollary}

\begin{proof}
Put $\eta_\alpha\coloneqq1-\frac1\alpha\in[0,\frac12]$. Lemma~\ref{lem:keycap} with
$K=1$ and $s=\sqrt v$ gives
\[
 \E\,e^{\lambda ZA}\;\le\;\exp\big(2^{4-\eta_\alpha}\lambda^2v\big)
 \qquad\text{for all }|\lambda|\le\frac{3^{\eta_\alpha}}{4\sqrt v},
\]
and $2^{4-\eta_\alpha}\le16$ while $3^{\eta_\alpha}/4\ge\frac14$ throughout $\alpha\in[1,2]$. This
bound is therefore at least as strong as the conclusion of Lemma~\ref{lem:lckey} (a
smaller quadratic constant on a wider range of $\lambda$), and
Corollary~\ref{cor:lcpairing} and the proof of Theorem~\ref{thm:logconcave} use nothing
else about the covariate law.
\end{proof}

\section{Discussion and Open Problems}\label{sec:discussion}

Exponential concentration suffices to retain the robustness scaling
$\eps\sqrt{nd/(cQ)}$, under the additional capacity condition $Q\le61n$.
This condition arises from the linear branch of the pairing tail bound.
Without it, the law holds with an extra factor of order $\sqrt{n/Q}$
for $Q\ge n$ (Proposition~\ref{prop:lcbeyond}). For $Q>2^{28}n$, the
construction of Proposition~\ref{prop:lcburnin}, whose covariates are
normalized log-concave, matches this degraded level up to an absolute
factor (Remark~\ref{rem:lcsharp}). 

The results leave unresolved how far the capacity condition can be
relaxed while retaining the stated threshold $L^\star$.

Two further questions concern the dimension required by the
counterexample and extensions to heterogeneous or dependent observations.

\begin{enumerate}[leftmargin=1.8em,itemsep=2pt]

\item \emph{Counterexamples in moderate dimension.}
The construction uses $d=e^{\Theta(Q)}$, assigning a separate coordinate
to each candidate function. Does comparable degradation occur when
$d=\operatorname{poly}(n,Q)$? Resolving this would clarify whether the
obstruction persists when candidate functions must share coordinates.

\item \emph{Heterogeneity and dependent observations.}
The companion paper \citep{PeterModular} establishes robustness laws
under isoperimetric concentration for heterogeneous mixtures and for
stationary Markov and $\beta$-mixing covariates.
Can these results be extended when the component distributions or
stationary marginal laws satisfy only exponential concentration?
What capacity, sample-size, and mixing conditions would then be needed?

\end{enumerate}

\acks{%
The framework developed here builds on the master's thesis \citep{Peter} and on the work of
\citet{BS21}. The author received no third-party funding or support for this work, 
and declares no competing financial interests.}

\appendix

\section{Probabilistic Toolbox}\label{app:toolbox}

This appendix collects further isoperimetric examples (see Section~\ref{sec:prelim}) and the 
proofs of the probabilistic facts used in Section~\ref{sec:toolbox}.

\paragraph{Further isoperimetric examples.}
Besides the isotropic Gaussian of Section~\ref{sec:prelim}, the textbook cases include
the uniform law on the unit sphere, via L\'evy's classical spherical isoperimetric
inequality \citep{Ledoux01}; and the uniform law on the solid cube rescaled to unit
diameter, via tensorization of the one-dimensional log-Sobolev inequality (or, in the
second edition's route, as a Lipschitz push-forward of Gaussian concentration)
\citep[Theorem~5.2.10]{Ver26}. For the two further mechanisms---uniformly convex
potentials, via the Bobkov--Ledoux logarithmic Sobolev inequality and the Herbst
argument, and positively curved compact Riemannian manifolds, via Gromov's isoperimetric
comparison---see \citet[Proposition~3.1]{BL00}, \citet[Theorem~2.2]{Gromov86} and
\citet{Ledoux01}.

\paragraph{Proofs of the Section~\ref{sec:toolbox} statements.}
Facts~\ref{fact:hoeffding} and~\ref{fact:bernstein} are classical
and are used as cited. The remaining proofs follow.

\phantomsection\label{proof:azuma}
\begin{proof}[Proof of Lemma~\ref{lem:azuma}]
If $B=0$ then $\sum_iV_i=0$ a.s.\ and the bound is trivial. Similarly, the bound is trivial when $s=0$.
So assume $B>0$ and $s>0$. Fix $\lambda\in\R$
and write $S_k\coloneqq\sum_{i=1}^kV_i$ for $0\le k\le n$, so that $S_0=0$.

\emph{Step 1 (conditional Hoeffding bound).}
Fix $i$ and let $\mu_i(\cdot\,;\omega)$ be a regular
conditional distribution of $V_i$ given $\mathcal G_{i-1}$, which exists because $V_i$ is a real
random variable. For $\Prob$-a.e.\ $\omega$ the law $\mu_i(\cdot\,;\omega)$ is supported in
$[-B,B]$ and has mean $\E[V_i\,|\,\mathcal G_{i-1}](\omega)=0$; applying
Fact~\ref{fact:hoeffding} to this law, with $[a,b]=[-B,B]$ and hence $(b-a)^2/8=B^2/2$, gives
\begin{equation}\label{eq:azuma-cond}
 \E\big[e^{\lambda V_i}\,\big|\,\mathcal G_{i-1}\big]
 =\int e^{\lambda v}\,\mu_i(dv;\cdot)\;\le\;e^{\lambda^2B^2/2}
 \qquad\text{a.s., for every }\lambda\in\R .
\end{equation}

\emph{Step 2 (tower property).} We claim $\E e^{\lambda S_k}\le e^{k\lambda^2B^2/2}$ for
$0\le k\le n$, by induction on $k$; the case $k=0$ is an equality. Let $1\le k\le n$ and suppose
the claim holds for $k-1$. Since $V_i$ is $\mathcal G_i$-measurable and
$\mathcal G_i\subseteq\mathcal G_{k-1}$ for every $i\le k-1$, the partial sum $S_{k-1}$ is
$\mathcal G_{k-1}$-measurable, and $|S_{k-1}|\le nB$ makes $e^{\lambda S_{k-1}}$ a bounded, hence
integrable, random variable. The tower property followed by the pull-out property of conditional
expectation therefore gives
\begin{align*}
 \E e^{\lambda S_k}
 &\;=\;\E\Big[\E\big[e^{\lambda S_{k-1}}e^{\lambda V_k}\,\big|\,\mathcal G_{k-1}\big]\Big]
 =\E\Big[e^{\lambda S_{k-1}}\;\E\big[e^{\lambda V_k}\,\big|\,\mathcal G_{k-1}\big]\Big]
 \;\le\;e^{\lambda^2B^2/2}\,\E e^{\lambda S_{k-1}} \\
 &\;\le\;e^{k\lambda^2B^2/2},
\end{align*}
where the first inequality uses the conditional Hoeffding bound from Step 1, and the
second is the induction hypothesis. Taking $k=n$ yields
$\E e^{\lambda S_n}\le e^{n\lambda^2B^2/2}$.

\emph{Step 3 (Chernoff).} For $\lambda>0$, Markov's inequality applied to $e^{\lambda S_n}>0$ gives
\[
 \Prob\Big[\sum_{i=1}^nV_i\ge s\Big]=\Prob\big[e^{\lambda S_n}\ge e^{\lambda s}\big]
 \le e^{-\lambda s}\,\E e^{\lambda S_n}\le\exp\Big(-\lambda s+\tfrac{n\lambda^2B^2}{2}\Big).
\]
The exponent is minimized at $\lambda=s/(nB^2)$, which is admissible ($\lambda>0$) when $s>0$;
substituting gives exponent $-s^2/(nB^2)+s^2/(2nB^2)=-s^2/(2nB^2)$, i.e.\ the claim.
\end{proof}

\phantomsection\label{proof:cappedtail}
\begin{proof}[Proof of Lemma~\ref{lem:cappedtail}]
By Markov's inequality,
$\Prob[W\ge s]=\Prob[e^{\theta W}\ge e^{\theta s}]\le\exp(a\theta^2-\theta s)$ for every
$\theta\in[0,\theta_0]$. Taking $\theta=\theta_0$ gives the second bound. The
unconstrained minimizer $\theta^\ast=\frac{s}{2a}$ of the exponent is admissible
precisely when $s\le2a\theta_0$, and substituting it gives
$a(\theta^\ast)^2-\theta^\ast s=-\frac{s^2}{4a}$.
\end{proof}

\phantomsection\label{proof:subgsq}
\begin{proof}[Proof of Lemma~\ref{lem:subgsq}]
\begin{enumerate}[label=(\alph*)]
 \item With $\theta=1/(4v)$, the layer-cake formula gives
  \begin{equation*}
    \E e^{\theta A^2}=1+\int_0^\infty \theta e^{\theta s}\,\Prob[A^2\ge s]\,ds
    \le 1+2\theta\int_0^\infty e^{(\theta-\frac1{2v})s}\,ds
    = 1+\frac{2\theta}{\frac1{2v}-\theta}=3\,.
  \end{equation*}
\item
  We similarly bound the $2k$-th moment:
  \begin{align*}
    \E A^{2k}&=\int_0^\infty 2kt^{2k-1}\Prob[|A|\ge t]\,dt\le 4k\int_0^\infty
    t^{2k-1}e^{-t^2/(2v)}dt=4k\cdot\tfrac12(2v)^k\Gamma(k)\\
    &=2(2v)^kk!\,.
  \end{align*}
\item By monotone convergence, valid since every term is nonnegative for $\theta\ge0$,
  \begin{equation*}
    \E e^{\theta A^2}=1+\sum_{k\ge1}\frac{\theta^k\E A^{2k}}{k!}
    \le 1+2\sum_{k\ge1}(2\theta v)^k=1+\frac{4\theta v}{1-2\theta v}\,.\qedhere
  \end{equation*}
\end{enumerate}
\end{proof}

\phantomsection\label{proof:psimom}
\begin{proof}[Proof of Lemma~\ref{lem:psimom}]
Using the layer-cake formula again, with the substitution $u=(t/s)^{\alpha}$, gives
\begin{align*}
   \E|B|^{k}&=\int_0^\infty kt^{k-1}\Prob[|B|\ge t]\,dt
            \;\le\;2k\int_0^\infty t^{k-1}e^{-(t/s)^{\alpha}}dt
            \;=\;\frac{2k\,s^{k}}\alpha\int_0^\infty u^{k/\alpha-1}e^{-u}\,du \\
            &=\;2s^{k}\,\frac k\alpha\,\Gamma\!\Big(\frac k\alpha\Big)\,,
\end{align*}
which equals $2s^{k}\Gamma(k/\alpha+1)$. For the second inequality set
$\beta\coloneqq1/\alpha\in[\frac12,1]$ and write
$\beta k+1=\beta\,(k+1)+(1-\beta)\cdot1$. Since $\log\Gamma$ is convex on $(0,\infty)$, we get
\[
 \Gamma\left(\frac{k}{\alpha}+1\right) = \Gamma(\beta k+1)\;\le\;\Gamma(k+1)^{\beta}\,\Gamma(1)^{1-\beta}\;=\;(k!)^{1/\alpha}.
 \qedhere
\]
\end{proof}

\section{Non-vacuity and Measurability}\label{app:nonvac}

This appendix contains the proof of Proposition~\ref{prop:overparam} and the
measurability analysis deferred from Remark~\ref{rem:measurable}.

\begin{proof}[Proof of Proposition~\ref{prop:overparam}]
By Lemma~\ref{lem:trunc},
$\frac1n\sum_i(\tau f_w(x_i)-y_i)^2\le\frac1n\sum_i(f_w(x_i)-y_i)^2$ for every $w$, so it
suffices to rule out below-noise fits by the truncated class. Since the pairs
$(x_i,y_i)$ are i.i.d., with $\cG_0\coloneqq\sigma(x_1,\dots,x_n)$ and
$\cG_i\coloneqq\sigma(\cG_0,z_1,\dots,z_i)$ we have
$\E[z_i\,|\,\cG_{i-1}]=\E[z_i\,|\,x_i]=0$ and $|z_i|\le2$. Define the events
\[
 G_1\coloneqq\Big\{\frac1n\sum_iz_i^2\ge\sigma^2-\frac{\eps}{8}\Big\},\qquad
 G_3\coloneqq\Big\{\sum_iz_i\big(-g(x_i)\big)\le\frac{n\eps}{16}\Big\}.
\]
Lemma~\ref{lem:noise} with $\xi=\eps/2\in(0,1]$ gives
$\Prob[G_1^c]\le e^{-n\eps^2/576}$, and Lemma~\ref{lem:globalsums}(ii) (with $K=1$ and
the filtration above) gives $\Prob[G_3^c]\le e^{-n\eps^2/2048}$.

Let $\cN\coloneqq\cN_\infty^{\cW}$ be the parameter net of Lemma~\ref{lem:net} with $L=\infty$. For
fixed $w'\in\cN$, the variables $V_i\coloneqq z_i\,\tau f_{w'}(x_i)$ are
$\cG_i$-measurable ($\tau\circ f_{w'}$ is Borel, as $f_{w'}$ is by hypothesis) with
$|V_i|\le2$ and
$\E[V_i\,|\,\cG_{i-1}]=\tau f_{w'}(x_i)\,\E[z_i\,|\,\cG_{i-1}]=0$, since
$\tau f_{w'}(x_i)$ is $\cG_0$-measurable; Lemma~\ref{lem:azuma} with $B=2$ and
$s=\frac{n\eps}{8}$ gives
\[
 \Prob\Big[\sum_iz_i\,\tau f_{w'}(x_i)\ge\frac{n\eps}{8}\Big]
 \;\le\;\exp\Big(-\frac{(n\eps/8)^2}{8n}\Big)\;=\;e^{-n\eps^2/512}.
\]
Summing over $\cN$ and applying Corollary~\ref{cor:netsum} with $E=n\eps^2/512$ --
applicable, since $n\eps^2\ge1024\,Q$ gives $E\ge2Q$ --
\[
 \Prob\Big[\exists w'\in\cN:\ \sum_iz_i\,\tau f_{w'}(x_i)\ge\frac{n\eps}{8}\Big]
 \;\le\;|\cN|\,e^{-n\eps^2/512}
 \;\le\;e^{-(E-Q)}\,\frac{\delta}{16}
 \;\le\;e^{-n\eps^2/1024}.
\]

On $H\coloneqq G_1\cap G_3\cap
\bigcap_{w'\in\cN}\big\{\sum_iz_i\,\tau f_{w'}(x_i)<\frac{n\eps}{8}\big\}$, no $w\in\cW$
fits below $\sigma^2-\eps$. Indeed, suppose
$\frac1n\sum_i(\tau f_w(x_i)-y_i)^2\le\sigma^2-\eps$ for some $w\in\cW$. By
Lemma~\ref{lem:net} there is $w'\in\cN$ with
$\frac1n\sum_i(\tau f_{w'}(x_i)-y_i)^2\le\sigma^2-\eps/2$; by Lemma~\ref{lem:decomp}
with $\xi=\eps/2$ and $f=\tau f_{w'}$---its first hypothesis holds on $G_1$---$\sum_iz_i\big(\tau f_{w'}(x_i)-g(x_i)\big)\ge\frac38\,n\,\frac\eps2=\frac{3}{16}n\eps$;
and on $G_3$, $\sum_iz_ig(x_i)\ge-\frac{n\eps}{16}$, whence
$\sum_iz_i\,\tau f_{w'}(x_i)\ge\frac{3}{16}n\eps-\frac{1}{16}n\eps=\frac{n\eps}{8}$,
contradicting membership in $H$. Finally
\[
 \Prob[H^c]\;\le\;e^{-n\eps^2/576}+e^{-n\eps^2/2048}+e^{-n\eps^2/1024}
 \;\le\;3e^{-n\eps^2/2048}.
\]
Since $H$ is a finite intersection of measurable events, the conclusion holds in the
sense of Remark~\ref{rem:measurable}.
\end{proof}

\paragraph{When the literal uniform event is measurable.}
Throughout, $a\coloneqq\sigma^2-\eps$, $R_f\coloneqq n^{-1}\sum_i(f(x_i)-y_i)^2$ and
$\mathcal U$ are the quantities of Remark~\ref{rem:measurable}.
Two comments on when $\mathcal U$ is genuinely measurable. First, the natural reduction
to a countable subclass stalls exactly at the fit threshold. Writing
$\mathcal U^{\mathsf c}=\bigcup_{w:\,\Lip(f_w)<L^\star}\{R_{f_w}\le a\}$, the
$J$-Lipschitz parametrization makes
$\cF_{<L^\star}\coloneqq\{f_w:\Lip(f_w)<L^\star\}$ separable in sup norm. Fix a
countable dense $\cG\subseteq\cF_{<L^\star}$. On a fixed sample, $f\mapsto R_f$ is
continuous in sup norm, since with $r\coloneqq\norm{h-f}_\infty$
\[
 |R_h-R_f|\;=\;\Big|\frac1n\sum_i\big(h(x_i)-f(x_i)\big)\big(h(x_i)+f(x_i)-2y_i\big)\Big|
 \;\le\;r^2+\frac{2r}{n}\sum_i|f(x_i)-y_i| ,
\]
so the infimum
\[
 Z\coloneqq\inf_{f\in\cF_{<L^\star}}R_f
   =\inf_{g\in\cG}R_g
\]
is a countable infimum of continuous functions of the sample (each $g\in\cG$ is
Lipschitz, hence continuous) and therefore Borel measurable. The non-strict threshold,
however, yields only the sandwich
$\{Z<a\}\subseteq\mathcal U^{\mathsf c}\subseteq\{Z\le a\}$ and the boundary
event $\{Z=a\}$ is not resolved. Under the strict-threshold hypothesis
$R_{f_w}<a$, i.e.\ for
\[
 \mathcal U_{<}\coloneqq\bigcap_{w\in\cW}
 \{R_{f_w}< a\Longrightarrow\Lip(f_w)\ge L^\star\}\,,
\]
the reduction closes: an infimum is below $a$ if and only if some member is, so
$\mathcal U_{<}^{\mathsf c}=\bigcup_{w:\,\Lip(f_w)<L^\star}\{R_{f_w}<a\}=\{Z<a\}$, and
$\mathcal U_{<}$ is Borel measurable outright (cf.\ \citealp[Remark~4.2]{Peter}). We keep
the non-strict hypothesis, which is the stronger statement and the form of
\citep[Theorem~3]{BS21}. Second, $\mathcal U$ is Borel whenever
$\cW$ is $\sigma$-compact (in particular, compact). Indeed $w\mapsto\Lip(f_w)$ is a
supremum of difference quotients, each continuous in $w$ by the $J$-Lipschitz
parametrization, hence lower semicontinuous, so
$\{w:\Lip(f_w)<L^\star\}=\bigcup_m\{w:\Lip(f_w)\le L^\star-\tfrac1m\}$ is $\sigma$-compact;
on each compact piece $K$ the continuous function $w\mapsto R_{f_w}$ attains its infimum,
which equals the infimum over a countable dense subset of $K$ and is therefore a
measurable function of the sample. Hence
$\mathcal U^{\mathsf c}=\bigcup_K\{\min_{w\in K}R_{f_w}\le a\}$ is a countable union of
measurable events.

\section{Functional Inequalities and Log-Concave Details}\label{app:lc}

This appendix collects the functional-inequality material deferred from
Section~\ref{sec:logconcave}: the product-Laplace counterexample separating exponential
concentration from isoperimetry, the proofs of Proposition~\ref{prop:poincexp} 
and Corollary~\ref{cor:logconcave}, and the subclass
catalog behind Remark~\ref{rem:nologd}.

\begin{proposition}[Exponential concentration is strictly weaker]\label{rem:lcnoconverse}
Exponential concentration does not imply $c$-isoperimetry for any $c$, even under the
covariance normalization of Corollary~\ref{cor:logconcave}: a product Laplace law has a
dimension-free Poincar\'e constant, hence dimension-free exponential concentration, but
for each fixed $d$ it is not $c$-isoperimetric for \emph{any} finite $c$, in particular
for no dimension-free one.
\end{proposition}
\begin{proof}
Let
$\mu=\operatorname{Lap}(0,d^{-1/2})^{\otimes d}$, where the second
argument denotes the Laplace scale parameter. Then
$\mathrm{Cov}(\mu)=(2/d)\Id_d$, so $\mu$ has the covariance normalization
of Corollary~\ref{cor:logconcave} (with $\kappa=2$). The one-dimensional measure
$\operatorname{Lap}(0,s)$ satisfies $\Var_{\nu_s}[h]\le4s^2\int|h'|^2\,d\nu_s$ for
bounded Lipschitz $h$ (\citealp[Lemma~2.1]{BL97}, by scaling); applying the Efron--Stein
inequality \citep[Theorem~3.1]{BLM13} and then this one-dimensional inequality
conditionally in each coordinate, as in the proof of
Proposition~\ref{prop:lcburnin}, gives
\[
 \Var_\mu[g]\le \frac4d\,\E_\mu\norm{\nabla g}^2
\]
for bounded Lipschitz $g$, which is the class for which
Proposition~\ref{prop:poincexp}'s proof invokes the hypothesis; $\mu$ is absolutely
continuous, so this Euclidean-gradient inequality yields the slope form over the same
class (see the paragraph following that proposition).
That proof therefore runs with $c_0=4$, so $\mu$ satisfies $36$-exponential
concentration, with a constant free of $d$.

On the other hand, fix $R>0$ and let
$f_R(x)\coloneqq\max\{-R,\min\{x_1,R\}\}$, which is bounded,
$1$-Lipschitz and, being odd in $x_1$ under the symmetric $\mu$,
centered. For $0\le t\le R$ one has $\{|f_R|\ge t\}=\{|X_1|\ge t\}$, an
event of probability $e^{-\sqrt d\,t}$ since $X_1\sim\operatorname{Lap}(0,d^{-1/2})$,
so $c$-isoperimetry would force
\[
 e^{-\sqrt d\,t}
 \;=\;\Prob\bigl[|f_R(X)-\E_\mu f_R|\ge t\bigr]
 \;\overset{\eqref{eq:isop}}{\le}\;2\exp\!\left(-\frac{dt^2}{2c}\right),
 \qquad 0\le t\le R,
\]
that is, $e^{-u}\le2e^{-u^2/(2c)}$ in the variable $u\coloneqq\sqrt d\,t$:
an exponential tail dominated by a Gaussian one. This fails at
$u=2c+2$, where
$\frac{u^2}{2c}-u=\frac{u(u-2c)}{2c}
=2+\frac2{c}>\log2$,
so choosing $R>(2c+2)/\sqrt d$ contradicts \eqref{eq:isop}. As
$c>0$ was
arbitrary, exponential concentration does not imply $c$-isoperimetry for any
$c$.
\end{proof}

\begin{proof}[Proof of Proposition~\ref{prop:poincexp}]
If $c_0=0$ the hypothesis forces $\Var_\mu[g]=0$ for every bounded Lipschitz $g$, so
$\mu$ is a Dirac mass and the conclusion is trivial; assume $c_0>0$. By rescaling
(replace $f$ by $(f-\E_\mu f)/L$ with $L\coloneqq\Lip(f)>0$ and evaluate the tail at
$t/L$; a constant $f$ makes the tail vanish) it suffices to treat a bounded
$1$-Lipschitz $f$ with
$\E_\mu f=0$; write $C_P\coloneqq c_0/d$ and $h(\lambda)\coloneqq\log\E e^{\lambda f}$.
Since $f$ is $1$-Lipschitz, the definition of the slope gives $|\nabla f|(x)\le1$ for
every $x\in\R^d$; no differentiability is invoked.

For $\lambda\in\R$, the function $g\coloneqq e^{\lambda f/2}$ is bounded Lipschitz ($f$
is bounded and Lipschitz, and $t\mapsto e^{\lambda t/2}$ is Lipschitz on the range of
$f$); only such $g$ enter the hypothesis in this proof. For every $x$,
\[
 |\nabla g|(x)\;\le\;\frac{|\lambda|}{2}\,e^{\lambda f(x)/2}\,|\nabla f|(x)
 \;\le\;\frac{|\lambda|}{2}\,e^{\lambda f(x)/2},
\]
because $|e^{\lambda f(y)/2}-e^{\lambda f(x)/2}|\le\frac{|\lambda|}2
\max\big(e^{\lambda f(y)/2},e^{\lambda f(x)/2}\big)\,|f(y)-f(x)|$ by the mean value
theorem applied to $t\mapsto e^{\lambda t/2}$, and the maximum tends to
$e^{\lambda f(x)/2}$ as $y\to x$ by continuity of $f$.

Applying the Poincar\'e inequality to $g$, with
$\E_\mu\big[\,|\nabla g|^2\,\big]\le\frac{\lambda^2}4\,\E e^{\lambda f}$ from the slope
bound, gives
$\E e^{\lambda f}-(\E e^{\lambda f/2})^2\le\frac{C_P\lambda^2}4\E
e^{\lambda f}$, i.e.
$h(\lambda)\le2h(\lambda/2)-\log\big(1-\tfrac{C_P\lambda^2}4\big)$
whenever $C_P\lambda^2<4$. Iterating $K$ times,
\[
 h(\lambda)\;\le\;2^{K}h\big(\lambda2^{-K}\big)
 +\sum_{k=0}^{K-1}2^{k}\Big(-\log\big(1-x\,4^{-k}\big)\Big),
 \qquad x\coloneqq\tfrac{C_P\lambda^2}4 .
\]
Since $f$ is bounded and centered, expanding the exponential and using $\E f=0$ gives
$\E e^{uf}\le1+\tfrac{u^2}2\norm f_\infty^2e^{|u|\norm f_\infty}$, hence
$h(u)\le\tfrac{u^2}2\norm f_\infty^2e^{|u|\norm f_\infty}$ and
$2^Kh(\lambda2^{-K})\le\lambda^22^{-K}\tfrac{\norm f_\infty^2}2e^{|\lambda|\norm f_\infty}$ 
goes to $0$ as $K\to\infty$. For $0\le\lambda\le1/\sqrt{C_P}$ we have
$x\le\frac14$, hence $-\log(1-x4^{-k})\le\frac{x4^{-k}}{1-x4^{-k}}
\le\frac43x4^{-k}$. The sum is then at most
$\frac43x\sum_{k\ge0}2^{-k}=\frac83x=\frac23C_P\lambda^2$. Thus
\begin{equation}\label{eq:poincmgf}
  \E\,e^{\lambda f} =  e^{h(\lambda)}\;\le\;e^{2/3 C_P\lambda^2}\;\le\;e^{C_P\lambda^2}
 \qquad\text{for }|\lambda|\le1/\sqrt{C_P}
\end{equation}
(the negative range by applying the argument to $-f$).
Due to Inequality~\eqref{eq:poincmgf}, we can apply Lemma~\ref{lem:cappedtail}
with $a=C_P$ and $\theta_0=1/\sqrt{C_P}$, which yields the tail bounds $e^{-t^2/(4C_P)}$ for
$t\le2\sqrt{C_P}$, and $e^{1-t/\sqrt{C_P}}$ for every $t\ge0$.

To obtain the two-sided exponential bound, first suppose that $t\le 3\log 2\,\sqrt{C_P}$.
Then $2\exp\!\left(-\frac{t}{3\sqrt{C_P}}\right)\ge1$,
and therefore
\[
 \Prob\bigl[|f-\E f|\ge t\bigr]
 \le1
 \le
 2\exp\!\left(-\frac{t}{3\sqrt{C_P}}\right).
\]

The second branch of Lemma~\ref{lem:cappedtail} is unrestricted in the
deviation, and \eqref{eq:poincmgf} covers $-f$ as well as $f$, so it applies
to both $f-\E f$ and $\E f-f$. Hence we get by the union bound
\[
 \Prob\bigl[|f-\E f|\ge t\bigr]
 \le
 2\exp\!\left(1-\frac{t}{\sqrt{C_P}}\right).
\]
Now suppose that
$t>3\log 2\,\sqrt{C_P}$. Then $t>3\log 2\,\sqrt{C_P}>\frac32\sqrt{C_P}$, so
$\frac{2t}{3\sqrt{C_P}}>1$ and hence
$1-\frac{t}{\sqrt{C_P}}<-\frac{t}{3\sqrt{C_P}}$.
Consequently,
\[
 \Prob\bigl[|f-\E f|\ge t\bigr]
 \le
 2\exp\!\left(-\frac{t}{3\sqrt{C_P}}\right) = 2\exp\!\left(-\frac{\sqrt{d}\,t}{\sqrt{9c_0}}\right)\,,
\]
since \(C_P=c_0/d\). This completes the proof of \eqref{eq:expconc} with \(c=9c_0\).
\end{proof}

\begin{proof}[Proof of Corollary~\ref{cor:logconcave}]
The proof of part~(i) has two steps. First we reduce to the isotropic case via an
affine change of variables, expressing the Poincar\'e constant of $\mu$ in
terms of that of an isotropic log-concave measure. We then invoke
Klartag's estimate for the latter. Part~(ii) replaces both steps by one bound.

\emph{Step 1: affine reduction to the isotropic case.}
Since $\mu$ is full-dimensional, $\Sigma\coloneqq\mathrm{Cov}(\mu)$ is
positive definite; let $\Sigma^{1/2}$ be its symmetric positive-definite
square root and $m\coloneqq\E_\mu[x]$. Then $\mu$ is the law of
$m+\Sigma^{1/2}\xi$ with $\xi$ an isotropic random variable ($\E\xi=0$, $\mathrm{Cov}(\xi)=\Id_d$). 
Write $T(x)\coloneqq m+\Sigma^{1/2}x$, so that $\xi=(T^{-1})_*\mu$; 
the isotropic measure $\xi$ inherits log-concavity from $\mu$ by affine
invariance, immediate from the Minkowski-combination definition of
log-concave measures \citep[Definition~1.1]{Borell74}.

Let $g\colon\R^d\to\R$ be locally Lipschitz. Since $X\sim\mu$ has the same
law as $T\xi$, we have $g(X)\overset{d}{=}(g\circ T)(\xi)$, whence
$\Var_\mu[g]=\Var_\xi[g\circ T]$. Every log-concave probability measure
satisfies a Poincar\'e inequality \citep{Bobkov99}; Step~2 below supplies the quantitative
estimate on the constant $C_P(\xi)$ via \citep[Theorem~1.2]{Klartag23} (see
also the survey \citealp{LV24}). Applying that inequality for $\xi$,
\[
 \Var_\xi[h]\;\le\;C_P(\xi)\,\E_\xi\norm{\nabla h}^2
 \qquad(h\text{ locally Lipschitz}\footnotemark),
\]
\footnotetext{This is the class of \citet[Eq.~(1.3)]{Klartag23}.}

to $h=g\circ T$, and using the chain rule (valid Lebesgue-a.e.\ by Rademacher's
theorem; the invertible affine $T$ preserves Lebesgue-null sets and $\xi$ is absolutely
continuous, so the displayed identity holds $\xi$-a.e.) together with the symmetry of
$\Sigma^{1/2}$,
\[
 \nabla(g\circ T)(x)
 \;=\;\big(\Sigma^{1/2}\big)^{\top}\nabla g(Tx)
 \;=\;\Sigma^{1/2}\,\nabla g(Tx),
\]
we obtain
\[
 \Var_\mu[g]\;=\;\Var_\xi[g\circ T]\;\le\;
 C_P(\xi)\,\E_\xi\big\|\Sigma^{1/2}\nabla g(Tx)\big\|^2 .
\]
For every $v\in\R^d$, $\norm{\Sigma^{1/2}v}^2=v^{\top}\Sigma v
\le\norm\Sigma_{\mathrm{op}}\norm v^2$; taking $v=\nabla g(Tx)$ and undoing the change of
variables ($Tx\sim\mu$, so $\E_\xi\norm{\nabla g(Tx)}^2=\E_\mu\norm{\nabla g}^2$)
gives
\[
 \Var_\mu[g]\;\le\;\norm\Sigma_{\mathrm{op}}\,C_P(\xi)\,
 \E_\mu\norm{\nabla g}^2 .
\]
Thus $C_P(\mu)\le\norm\Sigma_{\mathrm{op}}\,C_P(\xi)$: the affine map $T$
enlarges the Poincar\'e constant by at most $\norm\Sigma_{\mathrm{op}}$.

\emph{Step 2: Klartag's bound on the isotropic constant.}
For a probability measure $\nu$ on $\R^d$ with log-concave density
$\rho_\nu$, its \emph{isoperimetric constant} $\psi_\nu$ is defined \citep[Section~1]{Klartag23} by
\[
 \frac{1}{\psi_\nu}\;\coloneqq\;\inf_{A}\,
 \frac{\int_{\partial A}\rho_\nu}{\min\{\nu(A),\,1-\nu(A)\}}\,,
\]
the infimum running over open sets $A\subseteq\R^d$ with smooth boundary
and $0<\nu(A)<1$; $\psi_\nu$ is finite and positive \citep{Bobkov99}, as
recorded in \citep{Klartag23}. Thus $\psi_\nu$ is the reciprocal of the
Cheeger isoperimetric ratio, normalized so that Cheeger's inequality reads
$C_P(\nu)\le4\psi_\nu^2$ and so that the Kannan--Lov\'asz--Simonovits 
(KLS) conjecture is the assertion
that $\psi_\nu$ is bounded independently of $d$. Following
\citep[Eq.~(1.5)]{Klartag23}, define the \emph{dimension-$d$ KLS constant}
\[
 \psi_d\;\coloneqq\;\sup_{\nu}\,\psi_\nu ,
\]
the supremum over all isotropic log-concave probability measures $\nu$ on
$\R^d$ (each such $\nu$ has $\mathrm{Cov}(\nu)=\Id_d$, hence is
full-dimensional and admits a density by \citealp[Theorem~1.1]{Borell74},
so $\psi_\nu$ is defined). 

By \citep[Theorem~1.2]{Klartag23}, $\psi_d\le C\sqrt{\log d}$ for an
absolute constant $C$. The Cheeger--Buser--Ledoux equivalence, see
\citep[Eq.~(1.4)]{Klartag23} for the two-sided form
$\frac14\le\psi_\nu^2/C_P(\nu)\le\pi$, turns the left inequality into a bound for $\xi$:
$C_P(\xi)\le4\psi_d^2\le C_{\mathrm{KLS}}\log d$ with $C_{\mathrm{KLS}}\coloneqq4C^2$.

\emph{Conclusion.}
Combining the affine reduction with Klartag's estimate and using
$\norm\Sigma_{\mathrm{op}}\le\kappa/d$,
\[
 \Var_\mu[g]\;\le\;\frac{\kappa}{d}\,C_{\mathrm{KLS}}\log d\;\E_\mu\norm{\nabla g}^2
 \;=\;\frac{c_0}{d}\,\E_\mu\norm{\nabla g}^2,\qquad
 c_0\coloneqq C_{\mathrm{KLS}}\,\kappa\log d,
\]
which is the $d$-normalized Poincar\'e inequality of
Proposition~\ref{prop:poincexp} in its Euclidean-gradient form; $\mu$ is full-dimensional
log-concave, hence admits a density \citep[Theorem~1.1]{Borell74} and is absolutely
continuous, so the slope form holds as well (see the paragraph following that
proposition). It follows from that proposition that $\mu$ satisfies $c$-exponential
concentration with $c\le9c_0=9C_{\mathrm{KLS}}\,\kappa\log d\eqqcolon C_0\,\kappa\log d$, and we obtain
the stated Lipschitz lower bound from
Theorem~\ref{thm:logconcave}.

\emph{Part~(ii).} Granting \citet[Theorem~1.1]{BKL26} in its equivalent form
\citep[Eq.~(6)]{BKL26}, $\mu$ satisfies
\[
 \Var_\mu[g]\;\le\;C_{\mathrm{BKL}}\norm\Sigma_{\mathrm{op}}\,\E_\mu\norm{\nabla g}^2
 \;\le\;\frac{C_{\mathrm{BKL}}\,\kappa}{d}\,\E_\mu\norm{\nabla g}^2
\]
for every locally Lipschitz $g\in L^2(\mu)$, in particular for every bounded Lipschitz $g$.
This is the $d$-normalized Poincar\'e inequality of Proposition~\ref{prop:poincexp} with
$c_0\coloneqq C_{\mathrm{BKL}}\,\kappa$, in its Euclidean-gradient form.
Since $\mu$ is absolutely continuous, Rademacher's theorem gives
$|\nabla g|=\norm{\nabla g}$ $\mu$-a.e. Hence this is also the local-slope
Poincar\'e inequality required by Proposition~\ref{prop:poincexp}. That
proposition gives $c$-exponential concentration with $c\le9c_0=9C_{\mathrm{BKL}}\kappa$, and
Theorem~\ref{thm:logconcave} gives the stated bound.
\end{proof}

\paragraph{Subclasses with explicit dimension-free constants (Remark~\ref{rem:nologd}).}
For several important subclasses a dimension-free Poincar\'e constant is known with an
explicit value, so Step~2 can use it in place of Klartag's estimate, without appeal to
part~(ii). By Step~1, $c_0=\kappa\,C_P(\xi)$, so the proof yields
$c\le9\,C_P(\xi)\,\kappa$ with no logarithmic loss. Examples include:
\begin{enumerate}
\item \emph{Products of one-dimensional log-concave measures}
(e.g.\ the uniform measure on a box, or any distribution with independent
log-concave coordinates). Tensorization gives
\[
  C_P(\xi)=\max_i C_P(\xi_i):
\]
the inequality $\le$ is
\citep[Proposition~4.3.1]{BGL14} (stability of the Poincar\'e inequality
under products), iterated over the $d$ factors, and the reverse inequality
is immediate by testing against functions of a single coordinate.
Tensorization alone does not make this dimension-free; the second
ingredient is the universal one-dimensional bound: each coordinate
$\xi_i$ of the isotropic representative is a one-dimensional log-concave
law of unit variance, for which
$\frac1{12\Var[\xi_i]}\le\lambda_1(\xi_i)\le\frac1{\Var[\xi_i]}$ by
\citep[Corollary~4.3]{Bobkov99}, i.e.\ $C_P(\xi_i)\le12$, whence
$C_P(\xi)\le12$ uniformly in $d$.

\item \emph{Strongly log-concave measures},
$\xi\propto e^{-V}$ with $V$ smooth and
\[
  \nabla^2V(x)\succeq \rho \Id_d
  \qquad\text{for all }x,
\]
where $\rho>0$ is an absolute constant.
The Bakry--\'Emery criterion yields $C_P(\xi)\le \frac1\rho$
\citep[Corollary~4.8.2]{BGL14}. The isotropic Gaussian corresponds to
$\rho=1$, for which this constant is optimal (as noted after
\citealp[Proposition~4.8.1]{BGL14}).

\item \emph{The uniform measure on a Euclidean ball}, and more generally on
any ellipsoid $E=m+AB_2^d$ with $A$ invertible; see \citep[Theorem~1]{Bobkov03}.
\end{enumerate}

\section{Necessity of the Exponential-Tail Burn-in}\label{app:lcburnin}

This appendix proves Proposition~\ref{prop:lcburnin}: an explicit product-Laplace
instance in which every hypothesis of Theorem~\ref{thm:logconcave} except the burn-in
holds, while a below-noise fit with Lipschitz constant below $L^\star$ exists with high
probability.

\begin{proof}[Proof of Proposition~\ref{prop:lcburnin}]
Set
\[
 M\coloneqq\big\lceil e^{2\vartheta n}\big\rceil,
 \qquad
 d\coloneqq M,
 \qquad
 s\coloneqq d^{-1/2}.
\]
Let $\nu_s$ be the centered Laplace measure on $\R$ with density
\[
 u\longmapsto \frac1{2s}e^{-|u|/s},
\]
and let $\mu\coloneqq\nu_s^{\otimes d}$. Let
$(x_i,y_i)_{i\le n}$ be i.i.d., with $x_i\sim\mu$ and $y_i$ uniform on
$\{-1,+1\}$, independently of $x_i$. Then
\[
 g(x)=\E[y\mid x]=0,
 \qquad
 z_i=y_i,
 \qquad
 \sigma^2=\E z_i^2=1,
\]
so Assumption~\ref{ass:model} holds.

Define
\[
 r\coloneqq1-\sqrt{1-\eps},
\]
so that $0<r\le\eps$ and $(1-r)^2=1-\eps$. With $\tau$ denoting the
truncation from Section~\ref{sec:skeleton}, set
\[
 f_j(x)
 \coloneqq
 r\,\tau\!\left(\frac{\sqrt d\,x^{(j)}}{\vartheta}\right),
 \qquad
 j=1,\dots,M.
\]
Since $\tau$ is $1$-Lipschitz,
\[
 \Lip(f_j)\le \frac{r\sqrt d}{\vartheta}
 \le \frac{\eps\sqrt d}{\vartheta}.
\]
Moreover, because $\tau(t)=\operatorname{sign}(t)$ for $|t|\ge1$,
\[
 f_j(x)=r\operatorname{sign}(x^{(j)})
 \qquad\text{whenever}\qquad
 |x^{(j)}|\ge \frac{\vartheta}{\sqrt d} =  \vartheta s.
\]

\emph{Parametrization.}
Let
\[
 \cW
 \coloneqq
 \left\{
 \frac{j-1}{M-1}:j=1,\dots,M
 \right\}
 \subset\R,
 \qquad
 f_{(j-1)/(M-1)}\coloneqq f_j \quad \text{for } j=1,\dots,M\,.
\]
Then $p=1$ and $\operatorname{diam}(\cW)=1$. If $w\ne w'$, then
$|w-w'|\ge(M-1)^{-1}$, whereas
$\|f_w-f_{w'}\|_\infty\le2r\le2$. Hence
\[
 \|f_w-f_{w'}\|_\infty
 \le
 2(M-1)|w-w'|,
\]
so $\cF$ admits a $(1,J,1)$-parametrization with
$J\coloneqq2(M-1)$.

\emph{Exponential concentration.}
The one-dimensional Laplace measure satisfies
\[
 \Var_{\nu_s}[h]
 \le
 4s^2\int |h'|^2\,d\nu_s
\]
for bounded Lipschitz $h$; see \citep[Lemma~2.1]{BL97}, stated there at unit scale, with
the general case by scaling. 
Applying the
Efron--Stein inequality \citep[Theorem~3.1]{BLM13} to
$g(X_1,\dots,X_d)$, with $X_j$ independent and distributed according
to $\nu_s$, and then applying the preceding one-dimensional Poincar\'e
inequality conditionally in each coordinate, gives
\[
 \begin{aligned}
 \Var_\mu[g]
 &\le
 \sum_{j=1}^d
 \E\!\left[\Var\!\left[g(X)\mid X^{(-j)}\right]\right] \\
 &\le
 4s^2\sum_{j=1}^d\E_\mu[(\partial_jg)^2]
 =
 4s^2\E_\mu\|\nabla g\|^2
 =
 \frac4d\,\E_\mu\|\nabla g\|^2,
 \end{aligned}
\]
where $X^{(-j)}$ denotes $X$ with its $j$-th coordinate removed; hence the conditional
variance is taken only over the $j$-th coordinate. The partial derivatives $\partial_jg$
exist Lebesgue-a.e.\ by Rademacher's theorem, hence $\mu$-a.e.\ since $\mu$ has a
density, so $\sum_j(\partial_jg)^2=\norm{\nabla g}^2$ holds $\mu$-a.e.\ and the
displayed identity of integrals follows.
Thus the normalized Poincar\'e inequality of Proposition~\ref{prop:poincexp} holds with $c_0=4$ for bounded
Lipschitz $g$; since $\mu$ is absolutely continuous, the slope form holds for the same
class (see the paragraph following that proposition), and only such $g$ enter its proof.
Thus $\mu$ satisfies $c$-exponential concentration with $c=9c_0=36$.

\emph{Complexity.}
Since $p=W=1$ and $J=2(M-1)$,
\[
 Q
 =
 \log\!\left(\frac{96(M-1)}{\eps}\right)
 +
 \log\frac{16}{\delta}.
\]
Because $96(M-1)\ge M$,
\[
 Q>\log M\ge2\vartheta n.
\]
The sample-size condition and $\eps\le1$ imply
\[
 n\ge2^{13},
 \qquad
 \log\frac1\eps\le\frac12\log n,
 \qquad
 \log\frac{16}{\delta}\le\frac{n}{2^{12}}.
\]
Using also $M-1\le e^{2\vartheta n}$,
\[
 \begin{aligned}
 Q
 &\le
 2\vartheta n+\log96+\frac12\log n+\log\frac{16}{\delta}\\
 &\le
 2\vartheta n+\frac{n}{2^{10}}+\frac n2+\frac{n}{2^{12}}
 \le
 (2\vartheta+1)n.
 \end{aligned}
\]

\emph{Existence of a below-noise fit.}
For $j=1,\dots,M$, define
\[
 H_j
 \coloneqq
 \bigcap_{i=1}^n
 \{y_i x_i^{(j)}\ge\vartheta s\}.
\]
On $H_j$,
\[
 f_j(x_i)=r\,y_i
 \qquad(i=1,\dots,n),
\]
and therefore
\[
 \frac1n\sum_{i=1}^n(f_j(x_i)-y_i)^2
 =
 (1-r)^2
 =
 1-\eps
 =
 \sigma^2-\eps.
\]

Conditioned on $y=(y_1,\dots,y_n)$, the events $H_1,\dots,H_M$ are
independent because each depends on a distinct coordinate family.
Furthermore, symmetry of $\nu_s$ and
$\nu_s([\vartheta s,\infty))=\frac12e^{-\vartheta}$ give
\[
 \Prob[H_j\mid y]
 =
 \left(\frac12e^{-\vartheta}\right)^n
 =
 e^{-n(\vartheta+\log2)}
 \eqqcolon q.
\]
Hence, using the total probability formula, 
\[
 \Prob\!\left[\bigcap_{j=1}^M H_j^c\right]
 =
 (1-q)^M
 \le
 e^{-Mq}.
\]
Since
\[
 Mq
 \ge
 e^{2\vartheta n}e^{-n(\vartheta+\log2)}
 =
 e^{n(\vartheta-\log2)},
\]
we obtain
\[
 \Prob\!\left[\bigcap_{j=1}^M H_j^c\right]
 \le
 \exp\!\big(-e^{n(\vartheta-\log2)}\big).
\]
Finally,
\[
 e^{n(\vartheta-\log2)}\ge n,
\]
and therefore
\[
 \exp\!\big(-e^{n(\vartheta-\log2)}\big)
 \le
 e^{-n}
 \le
 e^{-2^{12}\log(16/\delta)}
 \le
 \frac{\delta}{16}.
\]
This proves (ii).

\emph{Lipschitz level.}
Since $Q\le(2\vartheta+1)n\le3\vartheta n$,
\[
 \Lip(f_w)
 \le
 \frac{\eps\sqrt d}{\vartheta}
 \le
 \frac{3\eps\sqrt d\,n}{Q}.
\]
As $c=36$,
\[
 \frac{3\eps\sqrt d\,n}{Q}
 =
 18\,\eps
 \sqrt{\frac{nd}{cQ}}
 \sqrt{\frac nQ}.
\]
Moreover, $Q>2\vartheta n$ and $\vartheta\ge2^{27}$ imply
\[
 \sqrt{\frac nQ}
 <
 \frac1{\sqrt{2\vartheta}}
 \le
 2^{-14}.
\]
Thus
\[
 \Lip(f_w)
 <
 18\cdot2^{-14}\,
 \eps\sqrt{\frac{nd}{cQ}}
 <
 2^{-9}\,
 \eps\sqrt{\frac{nd}{cQ}}
 =
 L^\star\,.
\]
This proves (i).

All assumptions of Theorem~\ref{thm:logconcave} are therefore satisfied
except the burn-in $n\ge Q/61$. With probability at least
$1-\delta/16$, some $f_w$ fits below the noise level while
$\Lip(f_w)<L^\star$, so the conclusion of that theorem fails on an
event of probability at least $1-\delta/16>\delta$.
\end{proof}


\begin{thebibliography}{99}
\bibitem[Boucheron et~al.(2013)]{BLM13} S.~Boucheron, G.~Lugosi, P.~Massart.
\emph{Concentration Inequalities: A Nonasymptotic Theory of Independence.}
Oxford University Press, 2013.

\bibitem[Bakry et~al.(2014)]{BGL14} D.~Bakry, I.~Gentil, M.~Ledoux.
\emph{Analysis and Geometry of Markov Diffusion Operators.} Grundlehren der mathematischen
Wissenschaften~348, Springer, 2014.

\bibitem[Bizeul et~al.(2026)]{BKL26} P.~Bizeul, B.~Klartag, J.~Lehec.
Presenting a proof of the Kannan--Lov\'asz--Simonovits conjecture.
arXiv:2610.05474 (v1), 2026. Result and equation numbers cited follow v1.

\bibitem[Bobkov(1999)]{Bobkov99} S.G.~Bobkov.
Isoperimetric and analytic inequalities for log-concave probability measures.
\emph{Annals of Probability} 27(4):1903--1921, 1999.

\bibitem[Bobkov(2003)]{Bobkov03} S.G.~Bobkov.
Spectral gap and concentration for some spherically symmetric probability measures.
In \emph{Geometric Aspects of Functional Analysis: Israel Seminar 2001--2002},
V.D.~Milman and G.~Schechtman (eds.), Lecture Notes in Math.~1807, Springer,
Berlin, 2003, pp.~37--43.

\bibitem[Bobkov and Ledoux(1997)]{BL97} S.~Bobkov, M.~Ledoux.
Poincar\'e's inequalities and Talagrand's concentration phenomenon for the exponential
distribution. \emph{Probability Theory and Related Fields} 107(3):383--400, 1997.

\bibitem[Bobkov and Ledoux(2000)]{BL00} S.G.~Bobkov, M.~Ledoux.
From Brunn--Minkowski to Brascamp--Lieb and to logarithmic Sobolev inequalities.
\emph{Geometric and Functional Analysis (GAFA)} 10(5):1028--1052, 2000.

\bibitem[Borell(1974)]{Borell74} C.~Borell.
Convex measures on locally convex spaces.
\emph{Arkiv f\"or Matematik} 12(1--2):239--252, 1974.

\bibitem[Bombari et~al.(2023)]{BKM23} S.~Bombari, S.~Kiyani, M.~Mondelli.
Beyond the universal law of robustness: sharper laws for random features and neural tangent
kernels. \emph{Proc.\ ICML 2023}, PMLR 202:2738--2776. arXiv:2302.01629.

\bibitem[Bubeck et~al.(2021)]{BLN21} S.~Bubeck, Y.~Li, D.~Nagaraj.
A law of robustness for two-layers neural networks.
\emph{Proc.\ COLT 2021}, PMLR 134:804--820. arXiv:2009.14444.

\bibitem[Bubeck and Sellke(2021)]{BS21} S.~Bubeck, M.~Sellke.
A universal law of robustness via isoperimetry.
\emph{Advances in Neural Information Processing Systems 34 (NeurIPS 2021)}, pp.~28811--28822.


\bibitem[Bubeck and Sellke(2023)]{BS23} S.~Bubeck, M.~Sellke.
A universal law of robustness via isoperimetry.
\emph{Journal of the ACM} 70(2), Article 10, 18~pp., 2023. doi:10.1145/3578580.

\bibitem[Evans and Gariepy(2015)]{EvansGariepy15} L.~C.~Evans, R.~F.~Gariepy.
\emph{Measure Theory and Fine Properties of Functions.} Revised edition, Textbooks in
Mathematics, CRC Press, Boca Raton, FL, 2015.

\bibitem[Gromov and Milman(1983)]{GM83} M.~Gromov, V.~Milman.
A topological application of the isoperimetric inequality.
\emph{American Journal of Mathematics} 105(4):843--854, 1983.

\bibitem[Gromov(1986)]{Gromov86} M.~Gromov.
Isoperimetric inequalities in Riemannian manifolds. Appendix in V.~D. Milman,
G.~Schechtman, \emph{Asymptotic Theory of Finite Dimensional Normed Spaces}, Lecture
Notes in Math.~1200, Springer, Berlin, 1986, pp.~114--129.

\bibitem[Klartag(2023)]{Klartag23} B.~Klartag.
Logarithmic bounds for isoperimetry and slices of convex sets.
\emph{Ars Inveniendi Analytica} (2023), Paper No.~4, 17~pp. doi:10.15781/jsjy-0b06.
arXiv:2303.14938.

\bibitem[Kumar et~al.(2024)]{KLS24} T.~Kumar, K.~Luo, M.~Sellke.
No free prune: information-theoretic barriers to pruning at initialization.
\emph{Proc.\ ICML 2024}, PMLR 235:25662--25681. arXiv:2402.01089.

\bibitem[Lata\l a and Oleszkiewicz(2000)]{LO00} R.~Lata\l a, K.~Oleszkiewicz.
Between Sobolev and Poincar\'e. In \emph{Geometric Aspects of Functional Analysis},
Lecture Notes in Math.~1745, Springer, 2000, pp.~147--168.

\bibitem[Ledoux(2001)]{Ledoux01} M.~Ledoux.
\emph{The Concentration of Measure Phenomenon.} AMS Mathematical Surveys and Monographs~89,
2001.

\bibitem[Lee and Vempala(2024)]{LV24} Y.~T.~Lee, S.~Vempala.
Eldan's stochastic localization and the KLS conjecture: isoperimetry, concentration and mixing.
\emph{Annals of Mathematics} 199(3):1043--1092, 2024.

\bibitem[Peter(2022)]{Peter} T.~J.~Peter.
\emph{Concentration of Measure and a Law of Robustness in Machine Learning.}
Master's thesis in Data Science, ETH Z\"urich, July 2022.

\bibitem[Peter(2026)]{PeterModular} T.~J.~Peter.
A modular law of robustness: heterogeneous mixtures, vector-valued predictors, and
dependent data. Companion manuscript, 2026.

\bibitem[Shmalo(2026)]{Shmalo26} Y.~Shmalo.
A law of robustness for two-layer neural networks with arbitrary weights.
arXiv:2607.07778 (v1), 2026. Result numbers cited follow v1, the only version as of
August 2026.

\bibitem[Song and Zhang(2026)]{SongZhang26} Z.~Song, X.~Zhang.
An $O(1)$ bound for the KLS constant.
arXiv:2610.01447 (v2), 2026.

\bibitem[Vershynin(2026)]{Ver26} R.~Vershynin.
\emph{High-Dimensional Probability: An Introduction with Applications in Data Science.}
2nd ed., Cambridge University Press, 2026.

\bibitem[Wu et~al.(2023)]{WHZ23} Y.~Wu, H.~Huang, H.~Zhang.
A law of robustness beyond isoperimetry.
\emph{Proc.\ ICML 2023}, PMLR 202:37439--37455. arXiv:2202.11592. Result numbers cited follow
the PMLR version; the earlier arXiv version (v1, 2022, circulated as ``How many data are
needed for robust learning?'') numbers the formal packing bound Theorem~3.6 and contains
additional appendices -- in particular Theorem~D.4, cited in
Proposition~\ref{prop:overparam} and Remark~\ref{rem:nonvacuous}, appears in the arXiv
version only.
\end{thebibliography}
\end{document}